\RequirePackage{fix-cm}
\documentclass[smallcondensed]{svjour3}     
\smartqed  
\usepackage[utf8]{inputenc}
\usepackage[T1]{fontenc}
\usepackage{times}
\usepackage{graphicx} 
\usepackage[tight,footnotesize]{subfigure} 

\usepackage{algorithm}
\usepackage{algorithmic}

\usepackage[linesnumbered,ruled,algo2e]{algorithm2e}

\usepackage{float}
\usepackage{hyperref}

\usepackage{amsmath,amssymb,array,cancel,color,setspace,stfloats,tikz,url}
\usetikzlibrary{calc,fit}
\usepackage[group-separator={,},group-digits=true]{siunitx}
\usepackage{enumitem}
\setenumerate{itemsep=1pt,topsep=2pt}

\usepackage{natbib}

\newcommand{\ie}{\textit{i.e.}, }
\newcommand{\eg}{\textit{e.g.}, }

\def\|{\,|\,}

\def\dir#1{\mathop{\rm Dir}\left(#1\right)}

\def\obj{{\bf x}}
\def\eqref#1{Eq~\ref{#1}}

\def\P{{\rm P}}

\def\obj{{\bf x}}

\def\graph{{\mathcal{G}}}
\def\bn{{\mathcal{B}}}

\def\BN{{ \mathop{ \textrm{BN} } }}

\def\data{\ifmmode \mathcal D\else$\data$\fi}
\def\labels{\ifmmode \mathcal L\else$\labels$\fi}
\def\card{\ifmmode \mathcal X\else$\card$\fi}

\def\k{\ifmmode \|\!\mathcal{Y}\!\|\else$\k$\fi}

\def\train{\ifmmode \mathcal T\else$\train$\fi}
\def\test{\ifmmode \mathcal U\else$\train$\fi}
\def\model{\ifmmode \mathcal M\else$\model$\fi}

\def\P{{\rm P}}
\def\ANDE^#1{\mathop{{\rm A}#1{\rm DE}}}

\def\eqref#1{Eq~\ref{#1}}

\DeclareGraphicsExtensions{.pdf,.png,.gif,.jpg}
\graphicspath{{../}}

\newlength{\figwidth}
\definecolor{citecol}{rgb}{0,0,0.5}
\hypersetup{urlcolor=citecol,linkcolor=black,citecolor=citecol,colorlinks=true} 

\let\cite\citep

\newcommand{\modif}[1]{{#1}}
\newcommand{\modiff}[1]{{#1}}

\title{Accurate parameter estimation for Bayesian network classifiers using hierarchical Dirichlet processes}
\titlerunning{Accurate parameter estimation for BN classifiers using HDP}

\journalname{Machine Learning}

\begin{document}
\sloppy

\author{Fran\c{c}ois Petitjean \and Wray Buntine \and Geoffrey~I.~Webb \and Nayyar Zaidi}

\institute{
Fran\c{c}ois Petitjean\at
              Faculty of Information Technology, Monash University\\
              \email{francois.petitjean@monash.edu}
              \and
              Wray Buntine\at
              Faculty of Information Technology, Monash University\\
              \email{wray.buntine@monash.edu}
              \and
              Geoffrey I. Webb\at
              Faculty of Information Technology, Monash University\\
              \email{geoff.webb@monash.edu}
              \and
              Nayyar Zaidi\at
              Faculty of Information Technology, Monash University\\
              \email{nayyar.zaidi@monash.edu}
}

\date{Received: date / Accepted: date}
\maketitle

\begin{abstract}
This paper introduces a novel parameter estimation method for the probability tables of Bayesian network classifiers (BNCs), using hierarchical Dirichlet processes (HDPs). The main result of this paper is to show that improved parameter estimation allows BNCs to outperform leading learning methods such as Random Forest for both 0-1 loss and RMSE, albeit just on categorical datasets.
  
As data assets become larger, entering the hyped world of ``big'', efficient accurate classification requires three main elements: (1) classifiers with low-bias that can capture the fine-detail of large datasets (2) out-of-core learners that can learn from data without having to hold it all in main memory and (3) models that can classify new data very efficiently. 

The latest Bayesian network classifiers (BNCs) satisfy these requirements. Their bias can be controlled easily by increasing the number of parents of the nodes in the graph. Their structure can be learned out of core with a limited number of passes over the data. However, as the bias is made lower to accurately model classification tasks, so is the accuracy of their parameters' estimates, as each parameter is estimated from ever decreasing quantities of data. In this paper, we introduce the use of Hierarchical Dirichlet Processes for accurate BNC parameter estimation even with lower bias.

We conduct an extensive set of experiments on 68 standard datasets and demonstrate that our resulting classifiers perform very competitively with Random Forest in terms of prediction, while keeping the out-of-core capability and superior classification time.
\end{abstract}

\section{Introduction}
\label{sec:Introduction}
With the ever increasing availability of large datasets, 
Bayesian network classifiers (BNCs) show great potential because
they can be learned out-of-core, {i.e.}\ without
having to hold the data in main memory. This can be done
in a discriminative fashion, for example, TAN \cite{friedman:bnc}, kDB \cite{sahami:lldbc} and Selective kDB (SkDB) \cite{Martinez2016} as well as generatively, using fixed-structure models such as na\"{i}ve Bayes \cite{Lewis1998} and average
n-dependence estimators -- AnDE \cite{WebbBoughtonWang05,WebbEtAl12}.
In contrast, random forests (RFs) \cite{Breiman:RF},
are not easily learned out-of-core because
they require either repeated sorting of the datasets or sampling.
\modif{Standard implementations side-step the problem either by ensuring that the training sets for each tree of the forest is small enough to be in-core \cite{Mahout}, or by relying on on-disk operations \cite{XGBoost}.}

Constraints on the network structure of BNCs are usually considered to be the main control on their bias-variance trade-off. If the number of parents for nodes is restricted to a relatively low number, then bias will generally be high and the variance on their estimates relatively low (we will actually show in the experiments that the variance can be high even for structures with low complexity). For large datasets, lower bias or higher complexity is preferable because it allows the models to more precisely capture fine detail in the data, translating into higher accuracy (exemplified by the success of deep networks).
The number of parameters to estimate increases exponentially with the number of parents allowed for each node; thus, for larger models, accurate estimation of the parameters becomes critical.

We now turn to the aim of this current paper. One of the main issues
with low-bias learners is their variance; it is logical that when
increasing the number of free parameters, even with the largest
possible dataset, there will be a point at which some parameters
will not have sufficient examples to be learned with precision. Variance is thus
not just a problem for small datasets, but can reappear when designing
effective learners for large datasets because they require low bias.
When the number of examples per parameter decreases, the variance increases
because parameter estimation fails to derive accurate estimates.
This, of course, is why maximum-likelihood estimates (MLEs)
are not often used with low-bias learners unless ensembles are also involved.

Remarkably, experiments in this paper show that for networks as simple as TAN (where each node has two parents at most), which significantly underperform RFs when using Laplace smoothing, can significantly outperform RFs once more careful parameter estimation is performed. 
This is particularly surprising because one wouldn't expect the variance to be high for models such as TAN. 
This is due to the fact that the variance is not even among all combinations of feature values and can indeed be relatively high for some of them. 
We will see that our estimates automatically adapt to cases with high or low variance by careful use of the hierarchical Dirichlet process (HDP).

\emph{Drawing the link between BNCs and HDP:}
Say you want to estimate the cancer rate in a population and
you are only given 10 samples; you will get a very crude estimate.  In
effect, this happens 100's of times over at each leaf of a decision
tree or clique of a Bayesian network when data is not abundant at the node.
For n-gram models, where one wishes to estimate extremely low-bias categorical distributions and for which very few examples per parameter are available, MLEs have long since been abandoned in favour of sophisticated smoothing techniques such as modified Kneser-Ney \cite{Chen:1996}.
These, however, have complex back-off parameters that need to be set.
For our more general and heterogeneous context of probability table estimation, there exist no techniques to set these parameters.
Hierarchical Pitman-Yor process (HPYP) is the Bayesian version of Kneser-Ney smoothing; it was introduced by \citet{Teh2006a} and uses empirical estimates for hyperparameters. 
This has been demonstrated to be very effective \cite{WooGasArc2011a,ShareghiIJCAI17}.
HPYP is well-suited for Zipfian contexts:  where discrete variables have
hundreds or more outcomes with very biased probabilities. Since we have discrete variables with mostly fewer outcomes
we do not use the HPYP, and prefer the lower-variance hierarchical Dirichlet process (HDP) \cite{Teh2006} -- it is equivalent to HPYP with discount parameter fixed to 0. 

In this paper, we propose to adapt the method of \citet{Teh2006a} for parameter estimation for n-gram models
and apply it to parameter estimation for BNCs.
Rather than the HPYP used by \citet{Teh2006a} we use the more computationally efficient HDP.
In this context, the model is simpler because a HDP with a finite discrete
base distribution is by definition equivalent to a Dirichlet distribution,
that is HDPs become hierarchical Dirichlet distributions in our context.
While conceptually simpler, we still use HDP style algorithms,
albiet more recent collapsed techniques,
because they are relatively efficient compared to the
older Chinese restaurant style algorithms
\cite{Buntine:2014,LimIJAR16}.

Having shown that our approach outperforms state-of-the-art BNC parameter estimation techniques, we use RF as an exemplar of state-of-the-art machine
learning because it is a widely used learning method for the types of
tabular data to which our methods are suited which can be used out of the
box without need for configuration. We show that our estimator allows BNCs to
compete against RFs on categorical datasets.
Furthermore, because our method is completely out-of-core, we demonstrate that
we can obtain results on large datasets on standard computers with which RF cannot even be trained using standard packages such as Weka.
Our models can also classify orders of magnitude faster than RF.

This paper is organized as follows. In Section~\ref{sec:BNCs}, we review Bayesian network classifiers (BNCs). 
In Section~\ref{sec:why-HDPs} we motivate our use of hierarchical Dirichlet Processes (HDPs) for BNCs' parameter estimation. 
We present our method in Section~\ref{sec:HDPs} and related work in Section~\ref{sec:Related work}. We have conducted extensive experiments, reported in Section~\ref{sec:Experiments}.

\section{Standard Bayesian network classifiers}
\label{sec:BNCs}
\subsection{Notations}
\label{subsec:BNCs}
\modif{Let $\data = \{ \obj^{(1)},\cdots,\obj^{(N)} \}$ be a dataset with $N$ objects.
Each datum $\obj = \langle x_1,\cdots,x_n \rangle$ is described over random variable $X_1,\cdots,X_n$.}
The following framework can be found in texts on learning Bayesian networks, such as \cite{Koller2009}.
A $\BN$ $\bn = \langle \graph,\Theta \rangle$, is characterized by the
structure $\graph$ (a directed acyclic graph, where each vertex $i$ is associated to a random variable $X_i$), and parameters $\Theta$, that quantifies the
dependencies within the structure. The parameter object
$\Theta$, contains a set of parameters for each vertex in $\graph$:
$\theta_{x_i|\Pi_i(\obj)}$, where $\Pi_i(.)$ is a function which given
the datum $\obj = \langle x_1,\ldots,x_n \rangle$ as its input,
returns the values of the attributes which are the parents of node $i$
in structure $\graph$.  Note, each attribute is a random variable
$X_i$ and $x_i$ represents the value of that random variable.  For
notational simplicity we write $\theta_{x_i|\Pi_i(\obj)}$ instead of
$\theta_{X_i = x_i | \Pi_i(\obj)}$.
We also use
$\theta_{X_i | \Pi_i(\obj)}$ to represent the full vector of values for
each $x_i$.
A $\BN$ $\bn$
computes the joint probability distribution as
\begin{equation*}
\P_\bn(\obj) = \prod_{i=1}^{n} \theta_{x_i | \Pi_i(\obj)}. 
\end{equation*}

\modif{The goal of developing a $\BN$ classifier is to predict the value of an additional variable $X_0=Y$: $X_0$ is the random variable associated with the class and we also denote it by $Y$ and its values by $y\in \mathcal{Y}$.
The data then takes the form $\data = \{ (y^{(1)},\obj^{(1)}),\ldots, (y^{(N)},\obj^{(N)}) \}$, the network takes an additional node and we can write:}
\begin{equation*} \label{eq_cl}
  \P_\bn(y|\obj) =  \frac{\P_\bn(y,\obj)}{\P_\bn(\obj)}
  = \frac{ \theta_{y |\Pi_0(\obj)} \prod_{i=1}^{n} \theta_{x_i | y, \Pi_i(\obj)} } {\sum_{y'\in \mathcal{Y}} \theta_{y' | \Pi_0(\obj)} \prod_{i=1}^{n} \theta_{x_i | y', \Pi_i(\obj)}}.
\end{equation*}
\modif{For simplicity, in the following, we use $\theta_y$ to denote $\theta_{y|\Pi_0(\obj)}$.}
Most notations are summarised in Table~\ref{tab:notations}.

\subsection{Structure learning for BNCs}\label{sec:structure-learning}
Most approaches to learning BNCs learn the structure first and then learn the parameters as a separate step. Numerous algorithms have been developed for learning BNC network structure. The key difference that distinguishes BNC structure learning  from normal BN structure learning is that the precision of the posterior estimates $\P_\bn(y|\obj)$  matters rather than the precision of $\P_\bn(y,\obj)$. As a result, it is usually important to ensure that all attributes in the class' Markov blanket are connected directly to the class or its children. As a consequence, it is common for BNCs to connect all attributes to the class.
\begin{figure}
        \centering
        \subfigure[\label{fig:kDB-1-structure}]{
        \begin{tikzpicture}
        [
        observed/.style={minimum size=10pt,circle,draw=blue!50,fill=blue!20},
        unobserved/.style={minimum size=10pt,circle,draw},
        hyper/.style={minimum size=1pt,circle,fill=black},
        post/.style={->,>=stealth',semithick},
        ]
        \node (x1) [observed] at (1,0) {\small $X_2$};    
        \node (x2) [observed] at (2.5,0) {\small $X_4$};
        \node (x3) [observed] at (4,0) {\small $X_1$};
        \node (x4) [observed] at (5.5,0) {\small $X_3$}; 
        \node (y) [observed] at (3.25,1.5) {\small $Y$};
        \draw[thick,->] (y) -- (x1);
        \draw[thick,->] (y) -- (x2);
        \draw[thick,->] (y) -- (x3);
        \draw[thick,->] (y) -- (x4);
        \draw[->,opacity=0.0]  (1,-0.55)--(5.5,-0.55) node[below,midway]{\small Decreasing mutual information with $Y$};
          \end{tikzpicture}
        }
        \subfigure[]{
    \begin{tikzpicture}
        [
        observed/.style={minimum size=12pt,circle,draw=blue!50,fill=blue!20},
        unobserved/.style={minimum size=12pt,circle,draw},
        hyper/.style={minimum size=1pt,circle,fill=black},
        post/.style={->,>=stealth',semithick},
        ]
        \node (x1) [observed] at (1,0) {\small $X_2$};    
        \node (x2) [observed] at (2.5,0) {\small $X_4$};
        \node (x3) [observed] at (4,0) {\small $X_1$};
        \node (x4) [observed] at (5.5,0) {\small $X_3$}; 
        \node (y) [observed] at (3.25,1.5) {\small $Y$};
        \draw[thick,->] (y) -- (x1);
        \draw[thick,->] (y) -- (x2);
        \draw[thick,->] (y) -- (x3);
        \draw[thick,->] (y) -- (x4);
        \path[thick,->] (x1) edge (x2);
        \path[thick,->] (x1) edge [bend left] (x3);		
        \path[thick,->] (x3) edge (x4);
        \draw[->]  (1,-0.55)--(5.5,-0.55) node[below,midway]{\small Decreasing mutual information with $Y$};
        \end{tikzpicture}
        }
        \caption{Example BNC structures\label{fig:structure}: (a) Na\"{i}ve Bayes, (b) kDB-1 }
\end{figure}
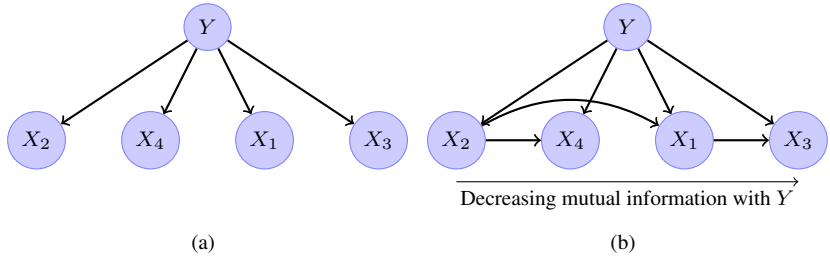
Na\"{i}ve Bayes (NB - see e.g. \cite{Lewis1998}) is a popular BNC that makes the class the parent of all other attributes and includes no other edges. The resulting network is illustrated in Figure~\ref{fig:structure}(a) and assumes conditional independence between all attributes conditioned on the class. As a consequence, 
$\P_\bn(y|\obj) \,\propto\,  \theta_{y} \prod_{i=1}^{n} \theta_{x_i | y} .$
Tree-augmented na\"{i}ve Bayes (TAN) \cite{friedman:bnc} adds a further parent to each non-class attribute, seeking to address the greatest conditional interdependencies. It uses the Chow-Liu \cite{Chow1968}
algorithm to find the maximum-likelihood tree of dependencies
among the attributes in polynomial time.

K-dependence Bayes (kDB) \cite{sahami:lldbc} allows each non-class attribute to have up to $k$ parents, with $k$ being a user-set value.
It first sorts the attributes on mutual information with the class. Each attribute $x_i$ is assigned the $k$ parent attributes that maximize conditional mutual information (CMI) with the class, $\textrm{CMI}(y,x_i|\Pi_i(\obj))$, out of those attributes with higher mutual information with the class. Figure~\ref{fig:structure}(b) shows kDB-1 (for $k=1$).

Selective kDB (SkDB) \cite{Martinez2016} selects values $n^*\leq n$ and $k^*\leq k$ such that a kDB over the $n^*$ attributes with highest mutual information with the class and using $k^*$ in place of $k$ maximizes some user selected measure of performance (in the current work, RMSE) assessed using incremental cross validation over the training data.

Other discriminative scoring schemes have been studied, see for example the work by \citet{carvalho2011discriminative}. A recent review of BNCs was written by \citet{bielza2014discrete}. 

\subsection{Maximum likelihood estimates}
\label{subsec:MLEs}
Given data points $\data = \{ (y^{(1)},\obj^{(1)}),\ldots, (y^{(N)},\obj^{(N)}) \}$, the log-likelihood of $\bn$ is:  
\begin{eqnarray}
  \label{eq_ll}
\sum_{j=1}^{N} \log \P_\bn(y^{(j)},\obj^{(j)})
= \sum_{j=1}^{N} \left( \log \theta_{y^{(j)} | \Pi_0(\obj^{(j)})} + \sum_{i=1}^{n} \log \theta_{X_i^{(j)} | y^{(j)},\Pi_i(\obj^{(j)})} \right),&&\\
 \label{eq_llconst}
\text{with }\sum_{y \in \mathcal{Y}} \theta_{y|\Pi_0(\obj)} = 1, \,\,\,\, \textrm{and}\, \sum_{X_i \in \card_i} \theta_{X_i|y,\Pi_i(\obj)} = 1.&&
\end{eqnarray}
Maximizing the log-likelihood to optimize the parameters ($\Theta$)
yields the well-known MLEs for Bayesian networks. 
Most importantly, MLEs factorize into independent distributions for each node, as do most standard maximum aposterior estimates \cite{Buntine.ieeekde}.
\begin{theorem} \label{th_ll} \cite{wermuth1982graphical}
Within the constraints in Equation~\ref{eq_llconst}, Equation~\ref{eq_ll} is maximized when $\theta_{x_i | \Pi_i(\obj)}$ corresponds to empirical estimates of probabilities from the data, that is, $\theta_{y | \Pi_0(\obj)} = \P_\data (y | \Pi_0(\obj))$ and $\theta_{X_i | \Pi_i(\obj)} = \P_\data (X_i | \Pi_i(\obj))$.
\end{theorem}
Thus our algorithms decompose the problem into separate sub-problems, one for each $ \theta_{X_i |y, \Pi_i(\obj)}$.

\subsection{Efficiency of BNC learning}
One often under-appreciated aspect of many BNC learning algorithms is their computational efficiency. Many BNC algorithms can be learned out-of-core, avoiding the overheads associated with retaining the training data in memory. 

NB requires only a single pass through the data to learn the parameters, counting the joint frequency of each pair of a class and an attribute value. 
TAN and kDB require two passes through the data. The first collects the statistics required to learn the structure, and the second the joint frequency statistics required to parameterize that structure.
SkDB requires three passes through the data. The first two collect the statistics required to learn structure and parameters, as per standard kDB. The third
performs an incremental cross validation to select a subset of the attributes and the $k^*\leq k$ to be used in place of $k$.

\section{Why and how are we using HDPs?}
\label{sec:why-HDPs}
The key contribution of this paper is to use hierarchical
Dirichlet processes for each categorical distribution $\theta_{X_i|\Pi_i(\obj)}$,
which yields back-off estimates that naturally smooth the
empirical estimates at the leaves.

The intuition for our method is that estimation of conditional probabilities should share information with their near neighbours. 
Suppose you wish to estimate a conditional probability table (CPT) for $\P(y|x_1,x_2,x_3)$ from data where the features $x_1,x_2,x_3$ take on values $\{1,2,3,4\}$.
This CPT can be represented as a tree: the root node branches on the values of $x_1$ and has 4 branches, the $2^{nd}$ and $3^{rd}$ level nodes test $x_2$ and $x_3$ and have 4 branches.
The $4^{th}$ level consists of leaves and each node has a probability vector for $y$ that we wish to estimate.
The sharing intuition says that the leaf node representing $\P(y|x_1=1,x_2=2,x_3=1)$ should have similar values to the leaf for
$\P(y|x_1=1,x_2=2,x_3=2)$ because they have a common parent,
but should not be so similar to $\P(y|x_1=3,x_2=1,x_3=2)$,
which only shares a great grandparent.

We achieve this sharing by using a hierarchical prior.
So we have vectors $\P(Y|x_1=1,x_2=2,x_3=u)$
(for $u=1,2,3,4$) that are generated from the same prior
with a common mean probability vector, say $q(Y|x_1=1,x_2=2)$.
Now $\P(y|x_1,x_2,x_3)$ can often be similar to $\P(y|x_1,x_2)$ which in turn can often be similar to $\P(y|x_1)$ and in turn to $\P(y)$. However, strictly speaking,
$\P(y|x_1,x_2)$, $\P(y|x_1)$ and $\P(y)$ are aggregate values here derived from the
underlying model which specifies $\P(y|x_1,x_2,x_2)$.
So, to model
hierarchical similarity with a HDP, instead of using
the derived $\P(y|x_1, x_2)$, $\P(y|x_1)$ and $\P(y)$ in the hierarchical prior,
we introduce some latent (hierarchical) parameters, say
$q(y|x_1,x_2)$, $q(y|x_1)$ and $q(y)$.  This indeed is the
innovation of \cite{Teh2006a}.   In our case we use
hierarchical Dirichlet distributions because the variables are
all discrete and finite, but the algorithm relies on methods
developed for a HDP \cite{LimIJAR16}.

\subsection{Intuition developed for na\"{i}ve Bayes}
Imagine a simple na\"{i}ve Bayes structure such as illustrated in Figure~\ref{fig:structure}(a): the class is the sole parent of every node in $\graph$. In this case, we use a (non-hierarchical) Dirichlet
as suggested for Bayesian na\"{i}ve Bayes \cite{Rennie:2003},
for $i=1,\cdots,n$ and all $y$
\begin{equation}\label{eq-NB-model}
\theta_{X_i|y}\sim\dir{\phi_{X_i},\alpha_{i}}~,
\end{equation}
where $\alpha_{i}$ is a (Dirichlet) concentration parameter for node $i$
(we will later develop how we \emph{tie} these parameters in
different configurations in the hierarchical case).
Note the non-standard notation for the Dirichlet:
for convenience we separate the vector probability $\phi_{X_i}$
and the concentration $\alpha_{i}$,
making it a 2-argument distribution.\footnote{Some papers would
  use the notation $\dir{\alpha_{i}\phi_{X_i}}$ or separate the vector
  $(\alpha_{i}\phi_{X_i})$ into its $|X_i|$ arguments.} 

We can think of this model in two ways:
we add a bias to the parameter estimation that encourages parameter estimates of each $\theta_{X_i|y}$ to have a common mean $\phi_{X_i}$ for different values of $y$.
Alternatively, we expect
$\theta_{X_i|y}$ for different values $y$ to be similar.
If they are similar, it is natural to think that they
have a common mean, in this case $\phi_{X_i}$.
Note, however, that $\phi_{X_i}$ is a {\it prior parameter},
introduced above as $q(\cdot)$,
and does not correspond to the mean estimated by marginalising with
$\sum_y \hat{p}(y)\theta_{X_i|y}$ readily estimated from the data.
The $\phi_{X_i}$ is a latent variable and a Bayesian hierarchical sampler is required to estimate it.

The hyperparameter $\alpha_{i}$,
called a concentration,
controls how similar the categorical distributions $\theta_{X_i|y}$ and $\phi_{X_i}$
should be: if $\alpha_{i}$ is large, then each $\theta_{X_i|y}$ virtually reproduces $\phi_{X_i}$; conversely, $\theta_{X_i|y}$ can vary more freely as $\alpha_{i}$ tends to 0.
Estimation also involves estimating the hyperparameters,
as discussed in Section~\ref{subsubsec:sampling concentration}.

\subsection{Intuition developed for \mbox{kDB-1}}
As described in Section~\ref{sec:structure-learning}, kDB-1 relaxes na\"{i}ve Bayes' assumption about the conditional independence (given $y$) between the
attributes by allowing one extra-parent per node
as presented in Figure~\ref{fig:kDB-1-structure}.
The structure learning process starts from the NB structure. Then it orders the nodes by highest mutual information with the class to be ranked first, \eg $\langle x_2,x_4,x_1,x_3\rangle$ in Figure~\ref{fig:kDB-1-structure}. Finally, it considers all candidate parents with higher mutual information with the class than itself (before in the order), and chooses the one that offers the highest mutual information between the class and the child node when conditioned on it. 
We keep the same idea for the estimation of $\theta_{X_i|\Pi_i(\obj)}$ as in the NB case, \eqref{eq-NB-model},
except that now $X_i$ has 2 parents: the class and another covariate. This translates into the following,
for $i=1,\cdots,n$ and all $y,\Pi{(i)}$
\begin{equation}
  \theta_{X_i|y,\Pi{(i)}}\sim\dir{\phi_{X_i|y},\alpha_{i|y}}~,
\end{equation}
where $\Pi(i)$ only comprises a single node for all $i>1$ (the first node has only $y$ as a parent).
Now we could have used $\phi_{X_i}$ as the latent parent,
so it is independent of $y$, but this would mean all leaves in the
tree have similar probability vectors.
This is a stronger statement than we need;  rather we prefer adjacent
nodes on the tree to be similar, not all nodes.
With a hierarchical model we add another level of complexity,
making the dependence on $y$ and require a further parent above
for $i=1,\cdots,n$ and all $y$
\begin{equation}
  \phi_{X_i|y}\sim\dir{\phi_{X_i},\alpha_{i|1}}~.
\end{equation}
This means that different branches in the tree can have
different means, and thus the model is more flexible (and has hence relatively low bias). 
Our Bayesian estimation handles these additional parameters and hyperparameters and limits the effect of variance on the model. 

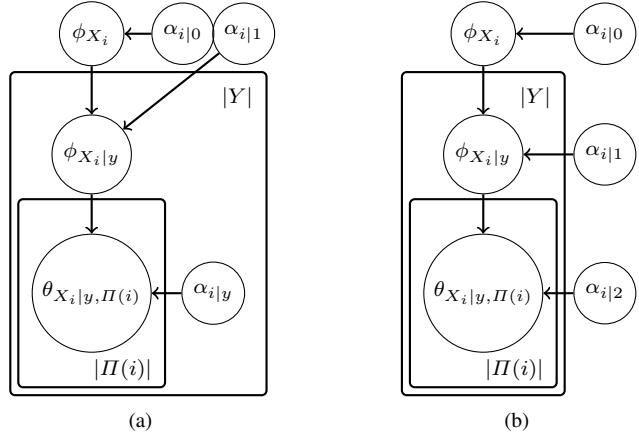
\begin{figure}
        \centering
        \subfigure[]{\small
        \begin{tikzpicture}
        [
        plate/.style={},scale=.8
    	]
       		\node[draw,circle] (xi) at (0,0) {$\phi_{X_i}$};
       		\node[draw,circle] (xigy) at (0,-2) {$\phi_{X_i|y}$};
            \node[draw,circle] (xigyp1) at (0,-4.3) {$\theta_{X_i|y,\Pi{(i)}}$};
            \node[draw,circle] (alpha0) at (1.5,0) {$\alpha_{i|0}$};
            \node[draw,circle] (alphai) at (2.5,0) {$\alpha_{i|1}$};
            \node[draw,circle] (alphaigy) at (2,-4.3) {$\alpha_{i|y}$};
            \draw[thick,->] (xi) -- (xigy);
			\draw[thick,->] (xigy) -- (xigyp1);
			\draw[thick,->] (alphai) -- (xigy);
            \draw[thick,->] (alpha0) -- (xi);
            \draw[thick,->] (alphaigy) -- (xigyp1);
            \node[draw, shape=rectangle, rounded corners=0.5ex, thick,align=right, inner sep=5pt, inner ysep=13pt,label={[xshift=-30pt,yshift=14pt]south east:$|\Pi{(i)}|$}, fit=(xigyp1)](plate1){};
            \node[draw, shape=rectangle, rounded corners=0.5ex, thick,align=right, inner sep=8pt, inner ysep=16pt,label={[xshift=-20pt,yshift=-17pt]north east:$|Y|$}, fit=(xigyp1) (xigy) (alphaigy)](plate2){};
        \end{tikzpicture}
        }\hspace{1.5cm}
        \subfigure[]{\small
        \begin{tikzpicture}
        [
        plate/.style={},scale=.8
    	]
       		\node[draw,circle] (xi) at (0,0) {$\phi_{X_i}$};
            \node[draw,circle] (alpha0) at (2,0) {$\alpha_{i|0}$};
       		\node[draw,circle] (xigy) at (0,-2) {$\phi_{X_i|y}$};
            \node[draw,circle] (xigyp1) at (0,-4.3) {$\theta_{X_i|y,\Pi{(i)}}$};
            \node[draw,circle] (alpha1) at (2,-2) {$\alpha_{i|1}$};
            \node[draw,circle] (alpha2) at (2,-4.3) {$\alpha_{i|2}$};
            \draw[thick,->] (xi) -- (xigy);
			\draw[thick,->] (xigy) -- (xigyp1);
			\draw[thick,->] (alpha1) -- (xigy);
            \draw[thick,->] (alpha0) -- (xi);
            \draw[thick,->] (alpha2) -- (xigyp1);
            \node[draw, shape=rectangle, rounded corners=0.5ex, thick,align=right, inner sep=5pt, inner ysep=13pt,label={[xshift=-30pt,yshift=14pt]south east:$|\Pi{(i)}|$}, fit=(xigyp1)](plate1){};
            \node[draw, shape=rectangle, rounded corners=0.5ex, thick,align=right, inner sep=8pt, inner ysep=16pt,label={[xshift=-20pt,yshift=-17pt]north east:$|Y|$}, fit=(xigyp1) (xigy)](plate2){};
        \end{tikzpicture}
        }
        \caption{\label{fig:kdb1-HDP}Our parameter structure model for one $X_i$ and kDB-1. (a) Tying the concentration at the parent. (b) Tying the concentration at the level. Details on tying are given in Section~\ref{subsubsec:sampling concentration}. }
\end{figure}

The model naturally defines the hierarchical structure given in Figure~\ref{fig:kdb1-HDP}, with the formula above represented by the graphical model given in Figure~\ref{fig:kdb1-HDP}(a).

\subsection{Intuition -- general framework}
The intuition of the framework for kDB-1 naturally extends to BNs with higher numbers of parents. 
We structure the estimation of the conditional probability of each factor ``child given parents'' to have a hierarchy with as many levels as the node has parents. 
At each level, the hierarchy branches on the different values that the newly introduced parent takes: on the different values of $y$ at the first level, on the different values of the first parent at the second level, etc. 
Once the structure is set, all we need is to have an order between the parents. 
For na\"{i}ve Bayes, there is only one parent -- $y$. 
For tree-augmented na\"{i}ve Bayes (TAN), as nodes cannot have more than a single parent apart from the class, we place the class first and its other parent second. 
For all other structures, we place $y$ as the first parent and then order the parents $\Pi_i$ by highest mutual information between them and the child conditioned on the class. 
This follows both the NLP framework for n-gram estimation and kDB structure learning:  position first in the hierarchy the nodes that are most likely to have an influence on the estimate. 
Positioning the class first allows us to pull the estimates to be most accurate in the probability space that is near $\P(y|\obj)$, which is our final target for classification, as we are not really interested in obtaining accurate estimates of $\P(X_i|y,\Pi(i))$ in parts of the probability space that are unrelated to $y$.

Note that the latent/prior probability vectors  $\phi_{X_i|y,\Pi_i(\obj)}$
do not model observed data, as the  $\theta_{X_i|y,\Pi_i(\obj)}$ do.
We represent them with different symbols
($\phi$ versus $\theta$) to highlight this fundamental difference.


Finally, note that in the finite discrete context, DPs are equivalent to Dirichlet
distributions \cite{Ferguson73}, so we present our models in terms of Dirichlets, but the inference is done efficiently using a collapsed Gibbs sampler for
HDPs \cite{du2010segmented,GasthausTeh10,Buntine:2014,LimIJAR16}.
These recent collapsed samplers for the hierarchical Bayesian algorithms are considerably more efficient and accurate and so do not suffer the well-known algorithmic issues of original hierarchical Chinese restaurant algorithms \cite{Teh2006}.
Note however that, unlike some applications of HDPs, there are no
`atoms' generated at the root of the HDP hierarchy
because the root is just a Dirichlet, which effectively
has the finite discrete set of atoms already present.
The HDP formalism is used to provide an efficient algorithm as a
collapsed version of a Gibbs sampler.

\section{Our framework: HDPs for BNCs}
\label{sec:HDPs}
This section reviews our model and sampling approach.

\subsection{Model}
\label{ssec:model}
\modiff{Consider the case of estimating $\P(X_c|y,x_1,\cdots,x_n)$ where $X_c$ represents the child variable of which we are trying to estimate the conditional probability distribution, and $y,x_1,\cdots,x_n$ are respectively used to denote the variable values $Y=y,X_1=v_1,\cdots,X_n=v_n$ . The variables $X_1,\cdots,X_n$ for $n\ge 0$ are ordered by mutual information with $X_c$ as described previously. Later, we will see that $X_c$ will represent the child variable in the Bayesian network of which we want to estimate the conditional probability distribution given its parents values $y,x_1,\cdots,x_n$. 
We can present this as a decision tree where the
root node banches on $y$ (\ie{}on the values of $Y$), all nodes at the $1^{st}$ level branch on $x_1$ (\ie{}on the values taken by $X_1$),
at the $2^{nd}$ level test $x_2$ and so forth.}
A node at the leaf (the $n+1$-th level) has the parameter vector
$\theta_{X_c|y,x_1,\cdots,x_n}$
for values of $y,x_1,\cdots,x_n$ given by its branch on the tree.
A node at the $i$-th level (for $i=1,\ldots,n$) has a parameter $\phi_{X_c|y,x_1,\cdots,x_i}$ --~which is a latent prior parameter~-- where again values of $y,x_1,\cdots,x_i$ are given by its branch on the tree.
The full hierarchical model is given by
\begin{eqnarray*}
  \theta_{\modiff{X_c}|y,x_1,\cdots,x_n} &\sim&\dir{\phi_{\modiff{X_c}|y,x_1,\cdots,x_{n-1}},\alpha_{y,x_1,\cdots,x_{n}}}\\
  \phi_{\modiff{\modiff{X_c}}|y,x_1,\cdots,x_i} &\sim&\dir{\phi_{\modiff{X_c}|y,x_1,\cdots,x_{i-1}},\alpha_{y,x_1,\cdots,x_{i}}}
  ~~~~~~~~~~~~~~\mbox{for }i=1,\ldots,n-1\\
  \phi_{\modiff{X_c}|y} &\sim&\dir{\phi_{\modiff{X_c}},\alpha_{y}}\\
  \phi_{\modiff{X_c}} &\sim&\dir{\frac{1}{|\modiff{X_c}|}\vec{1},\alpha_{0}} ~.
\end{eqnarray*}
\modiff{Note each Dirichlet has a concentration parameter as a hyperparameter,
and denote the full set of these by $\alpha_*$.
These are known to significantly change the characteristics of the
distribution, so they must be estimated as well.
We discuss below, in Section~\ref{subsubsec:sampling concentration}, how we can tie these hyperparameters $\alpha_*$ so that they are not all
distinct.} Experience has shown us that there should not be just one
value in the entire tree, nor should there be a different value \modiff{for each} node.

\subsection{\modif{Posterior inference}}
\label{subsec:Posterior inference}
\modif{To consider how posterior inference is done with this model,
first consider the simplest case of a single node with
probabilities
$\phi_{\modiff{X_c}|y}$ where
a data vector $n_{\modiff{X_c}|y}$ is sampled with total count
$n_{\cdot|y}\modiff{=\sum_{x_c}n_{x_c|y}}$:
\begin{eqnarray*}
  \phi_{\modiff{X_c}|y} &\sim&\dir{\phi_{\modiff{X_c}},\alpha_{y}}\\
  n_{\modiff{X_c}|y}&\sim&\mbox{multinomial}(\phi_{\modiff{X_c}|y},n_{\cdot|y}) ~.
\end{eqnarray*}
\modiff{For example, in Dataset~1 given later in Table~\ref{tab:worked-examples}, the values of $n_{x_1|y}$ are as follows, for each value of $X_1$ and $Y$:
  \begin{displaymath}
    \begin{array}{rcl}
      n_{0|0}&=&2\\
      n_{1|0}&=&0\\
      n_{0|1}&=&20\\
      n_{1|1}&=&5
    \end{array}
  \end{displaymath}
  These are contained into two vectors for $Y=0$ and $Y=1$:
  \begin{displaymath}
    \begin{array}{rcl}
      n_{X_1|0}&=&[2,0]\\
      n_{X_1|1}&=&[20,5]
    \end{array}
  \end{displaymath}
  The total count for the first vector are thus respectively $n_{.|0}=2$ and $n_{.|1}=25$.}
The marginalised likelihood for this,
which marginalises out $\phi_{\modiff{X_c}|y}$ takes the form \cite{Buntine.ieeekde}
\begin{equation}
  \label{eq-ms}
\P(n_{\modiff{X_c}|y}|\phi_{\modiff{X_c}},\alpha_{y},n_{\cdot|y})
=
{ n_{\cdot|y} \choose n_{\modiff{X_c}|y}}
  \frac{\Gamma(\alpha_{y})}{\prod_{\modiff{x_c}} \Gamma(\phi_{\modiff{x_c}|y}\alpha_{y})}
  \frac{\prod_{\modiff{x_c}} \Gamma(\phi_{\modiff{x_c}|y}\alpha_{y}+ n_{\modiff{x_c}|y})}{\Gamma(\alpha_{y}+n_{\cdot|y})} ~.
\end{equation}
\modiff{where $x_c$ represents the values taken by $X_c$.
Our goal in this is to estimate the $\phi_{\modiff{X_c}}$ parameters.
As it stands, this is going to be very costly because they appear in a
complex form inside gamma functions,
$\prod_{\modiff{x_c}}\frac{ \Gamma(\phi_{\modiff{x_c}|y}\alpha_{y}+ n_{\modiff{x_c}|y})}{\Gamma(\phi_{\modiff{x_c}|y}\alpha_{y})}$.
New collapsed methods developed for HDPs deal with this problem
by modifying it with the introduction of new (latent) variables that
make the gamma functions disappear.}

While one can formalise Equation~\ref{eq-ms} using HDPs,
in this case a direct augmentation can be done
using the identity (for $n\in {\cal N}^+$)
\begin{equation} \label{eq-si}
\frac{\Gamma(\alpha+n)}{\Gamma(\alpha)}
= \sum_{t=1}^n \alpha^t S^n_t
\end{equation}
where $S^n_t$ is an unsigned Stirling number of the first kind.
The Stirling number
is a combinatoric quantity that is easily tabulated \cite{du2010segmented}
and simple asymptotic formula exist \cite{HWANG1995}.}
\modiff{This is sometimes converted into the Chinese restaurant distribution (CRD)
in the form
\begin{equation} \label{eq-sicrd}
\P(t|CRD,n,\alpha) = \frac{\Gamma(\alpha)}{\Gamma(\alpha+n)} \alpha^t S^n_t
\end{equation}
and note the normalisation of Equation~\ref{eq-sicrd}
is shown by Equation~\ref{eq-si},
where $t\in\{1,...,n\}$ for $n>0$.}

\modiff{To simplify Equation~\ref{eq-ms}, multiply the LHS
by $\prod_\modiff{x_c} \P(t_{\modiff{x_c}|y}|CRD,n_{\modiff{x_c}|y},\phi_{\modiff{x_c}|y}\alpha_{y})$
and the RHS
by the corresponding RHSs from Equation~\ref{eq-sicrd}.
This is called an augmentation because we are 
are introducing new latent variables $t_{\modiff{x_c}|y}$
for each $\modiff{x_c}$,
represented in our notation as $t_{\modiff{X_c}|y}$.
The terms in $\Gamma(\phi_{\modiff{x_c}|y}\alpha_{y})$ etc.,
cancel out yielding}
\begin{eqnarray}
  \nonumber
  \P(n_{\modiff{X_c}|y},t_{\modiff{X_c}|y}|\phi_{\modiff{X_c}},\alpha_{y},n_{\cdot|y}) &=&
    { n_{\cdot|y} \choose n_{\modiff{X_c}|y}}
    \frac{\Gamma(\alpha_{y})}{\Gamma(\alpha_{y}+n_{\cdot|y})}
         \prod_\modiff{x_c} (\alpha_y\phi_{\modiff{x_c}})^{t_{\modiff{x_c}|y}} S^{n_{\modiff{x_c}|y}}_{t_{\modiff{x_c}|y}} \\
      \label{eq-msi}
      &=&
            { n_{\cdot|y} \choose n_{\modiff{X_c}|y}}
            \frac{\alpha_y^{t_{\cdot|y}}}{\alpha_y^{(n_{\cdot|y})}}
            \prod_\modiff{x_c} \phi_{\modiff{x_c}}^{t_{\modiff{x_c}|y}} S^{n_{\modiff{x_c}|y}}_{t_{\modiff{x_c}|y}}
\end{eqnarray}
where 
$\alpha^{(n)}=\alpha(\alpha+1)\cdots(\alpha+n-1)$ is a
rising factorial.

\noindent
\modiff{Notice what has been done here for the current nodes $\modiff{X_c}$:
\begin{itemize}
\item
  the parent probabilities $\phi_{\modiff{X_c}}$ now appear in a simple
  multinomial likelihood $\prod_\modiff{x_c} \phi_{\modiff{x_c}}^{t_{\modiff{x_c}|y}}$,
\item
  their prior complex form inside gamma functions has been eliminated,
\item
  but at the expense of introducing new latent variables $t_{\modiff{X_c}|y}$.
\end{itemize}
This operation forms the basis for simplifying a full tree of
such nodes recursively, presented in the
next section.
Equation~\ref{eq-msi} was originally developed and used in the
context of the HDP, but the above alternative derivation is
adequate for our purposes.}

\modiff{One can think of this in terms of Bayesian inference
on a DAG where evidence functions are passed between nodes.
Instead of passing the evidence represented by Equation~\ref{eq-ms}
from nodes $\modiff{X_c}$ to parent $y$, we pass the
evidence $\prod_\modiff{x_c} \phi_{\modiff{x_c}}^{t_{\modiff{x_c}|y}}$ which is
just a multinomial likelihood so it can be combined with the prior
in the usual manner.
So for every count $n_{\modiff{x_c}|y}>0$ in the tree, one is introducing
a {\em pseudo-count} $t_{\modiff{x_c}|y}$ as a latent variable,
where $1\leq t_{\modiff{x_c}|y} \leq n_{\modiff{x_c}|y}$.}

\paragraph{How does this relate to a Chinese restaurant process (CRP)?}
Suppose we have a Dirichlet process with base distribution
$\phi_\modiff{X_c}$ and we sample $n_{\cdot|y}$ data generating
a Chinese restaurant configuration,
where the $n_{\cdot|y}$ sample points are distributed over a number of tables.
Then the $t_{\modiff{x_c}|y}$ variables above corresponds to the number of tables
in the restaurant for data $\modiff{x_c}$, which is by definition
between 1 and $n_{\modiff{x_c}|y}$ when $n_{\modiff{x_c}|y}>0$ \cite{LimIJAR16}.
\modiff{Indeed the probability of the CRD above is the formula for a}
collapsed CRP \cite{du2010segmented,GasthausTeh10},
where the numbers of data at each table are marginalised out,
only keeping the count of tables.
This represents a huge advantage computationally
because one only needs to store the
number of tables at each node, not the full configuration
of customers at tables.  This eliminates the need for
dynamic memory that burdens a hierarchical CRP.

\subsection{Context tree -- data structure}
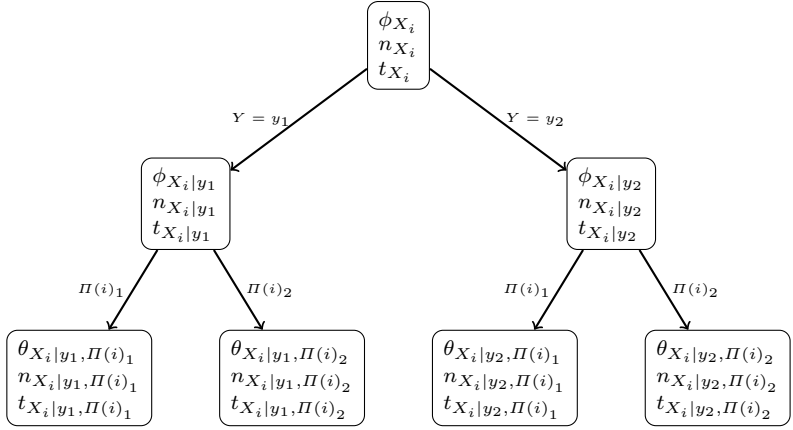
\begin{figure}
  \centering
  \small
  \begin{tikzpicture}
    [
    plate/.style={},scale=.7,rounded corners
    ]
    \node[draw,rectangle] (xi) at (0,0) {$\begin{array}{l}\phi_{X_i}\\n_{X_i}\\t_{X_i}\end{array}$};
    \node[draw,rectangle] (xigy1) at (-4,-3) {$\begin{array}{l}\phi_{X_i|y_1}\\n_{X_i|y_1}\\t_{X_i|y_1}\end{array}$};
    \node[draw,rectangle] (xigy2) at (4,-3) {$
      \begin{array}{l}
        \phi_{X_i|y_2}\\
        n_{X_i|y_2}\\
        t_{X_i|y_2}
      \end{array}
$};
    \node[draw,rectangle] (xigy1p1) at (-6,-6.3) {$
      \begin{array}{l}
        \theta_{X_i|y_1,\Pi{(i)}_1}\\
        n_{X_i|y_1,\Pi{(i)}_1}\\
        t_{X_i|y_1,\Pi{(i)}_1}
      \end{array}
      $};
    \node[draw,rectangle] (xigy1p2) at (-2,-6.3) {$
      \begin{array}{l}
        \theta_{X_i|y_1,\Pi{(i)}_2}\\
        n_{X_i|y_1,\Pi{(i)}_2}\\
        t_{X_i|y_1,\Pi{(i)}_2}
      \end{array}
      $};
    \node[draw,rectangle] (xigy2p1) at (2,-6.3) {$
      \begin{array}{l}
        \theta_{X_i|y_2,\Pi{(i)}_1}\\
        n_{X_i|y_2,\Pi{(i)}_1}\\
        t_{X_i|y_2,\Pi{(i)}_1}
      \end{array}
      $};
    \node[draw,rectangle] (xigy2p2) at (6,-6.3) {$
      \begin{array}{l}
        \theta_{X_i|y_2,\Pi{(i)}_2}\\
        n_{X_i|y_2,\Pi{(i)}_2}\\
        t_{X_i|y_2,\Pi{(i)}_2}
      \end{array}
      $};
    \path (xi) -- node[midway,left] (arrowxi1) {\tiny$Y=y_1$} (xigy1);
    \path (xi) -- node[midway,right] (arrowxi2) {\tiny$Y=y_2$} (xigy2);
    \path (xigy1) -- node[midway,left] (arrowxigy11) {\tiny$\Pi{(i)}_1$} (xigy1p1);
    \path (xigy1) -- node[midway,right] (arrowxigy12) {\tiny$\Pi{(i)}_2$} (xigy1p2);
    \path (xigy2) -- node[midway,left] (arrowxigy21) {\tiny$\Pi{(i)}_1$} (xigy2p1);
    \path (xigy2) -- node[midway,right] (arrowxigy22) {\tiny$\Pi{(i)}_2$} (xigy2p2);
    \draw[thick,->] (xi) -- (xigy1);
    \draw[thick,->] (xi) -- (xigy2);
    \draw[thick,->] (xigy1) -- (xigy1p1);
    \draw[thick,->] (xigy1) -- (xigy1p2);
    \draw[thick,->] (xigy2) -- (xigy2p1);
    \draw[thick,->] (xigy2) -- (xigy2p2);
  \end{tikzpicture}
  \caption{\label{fig:tree-kDB1}Context tree for our parameter structure model for kDB1 and one $X_i$.}
\end{figure}

The intuition of Equation \ref{eq-msi} is that
each node $\theta_{\modiff{X_c}|y,x_1,\cdots,x_n}$ or $\phi_{\modiff{X_c}|y,x_1,\cdots,x_i}$ passes up some fraction of its own data as a multinomial likelihood
to its parent. So the nodes will have a vector
of sufficient statistics $n_{\modiff{X_c}|y,x_1,\cdots,x_{i}}$ recorded for each node.
These have a virtual CRP with which
we only record the number of tables
$t_{\modiff{X_c}|y,x_1,\cdots,x_{i}}$,
which we refer to as {\em pseudo-counts}.
The counts $t_{\modiff{X_c}|y,x_1,\cdots,x_{i}}$ represents the
fraction of $n_{\modiff{X_c}|y,x_1,\cdots,x_{i}}$ that is passed
(in a multinomial likelihood) up to its parent node,
as dictated by Equation \ref{eq-msi}.
An example of context tree for kDB1 is given in Figure~\ref{fig:tree-kDB1}, which simply unfolds the plate notations used in Figure~\ref{fig:kdb1-HDP} and adds the $t$ and $n$ variables. 

As with hierarchical CRPs,
these statistics are related for $i\ge 0$:
\begin{equation}
\label{eq-crpstat}
n_{\modiff{x_c}|y,x_1,\cdots,x_{i-1}} = \sum_{x_i} t_{\modiff{x_c}|y,x_1,\cdots,x_{i}},
\end{equation}
and moreover the base case
$n_{\modiff{x_c}} = \sum_{y} t_{\modiff{x_c}|y}$.
\modiff{The counts $n_{\modiff{x_c}|y,x_1,\cdots,x_{i-1}}$
here only represent real data counts at the leaf nodes.
At internal nodes, the $n_*$ represent totals of psuedo-counts
from the child nodes, as passed up by the multinomial evidence messages
for the children.}

The likelihood for the data with this configuration
can be represented with $\theta$ and all  but the root
$\phi$ marginalised out:
\begin{equation}
\label{eq-dm}
\P(\data,n,t|\phi_{\modiff{X_c}},\alpha)=\left(\prod_{\modiff{x_c}} \phi_{\modiff{x_c}}^{n_\modiff{x_c}}\right)
  \prod_{i=0}^n \left(\prod_{y,x_1,\cdots,x_i} 
  \frac{\alpha_{y,x_1,\cdots,x_{i}}^{t_{\cdot|y,x_1,\cdots,x_{i}}}}{\alpha_{y,x_1,\cdots,x_{i}}^{(n_{\cdot|y,x_1,\cdots,x_{i}})}}
  \prod_{\modiff{x_c}} S^{n_{\modiff{x_c}|y,x_1,\cdots,x_{i}}}_{t_{\modiff{x_c}|y,x_1,\cdots,x_{i}}} \right)~,
\end{equation}
and the `dot' notation is used to represent totals, so $n_{\cdot|y}=\sum_\modiff{x_c} n_{\modiff{x_c}|y}$.
The multinomial likelihood on $\phi_{\modiff{X_c}}$  can also be marginalised
out with a Dirichlet prior.
Note the formula can be seen to be derived by recursive application
(bottom up) of the
formula in Equation~\ref{eq-msi}.

Once the parameters have been estimated (described in the next sub-section),
the parameters $\theta$ can be estimated recursively
using the standard hierarchical CRP estimation formula:
\begin{eqnarray}\label{eq-crpe1}
  \hat{\phi}_\modiff{x_c} &=& \frac{ n_\modiff{x_c} + \frac{1}{|\modiff{X_c}|}\alpha_0}{n_\cdot + \alpha_0}\\\label{eq-crpe2}
  \hat{\phi}_{\modiff{x_c}|y,x_1,\cdots,x_i} &=&
  \frac{ n_{\modiff{x_c}|y,x_1,\cdots,x_i} + \hat{\phi}_{\modiff{x_c}|y,x_1,\cdots,x_{i-1}}\alpha_{y,x_1,\cdots,x_i}  }
       {n_{\cdot|y,x_1,\cdots,x_i} +   \alpha_{y,x_1,\cdots,x_i}}   \\\label{eq-crpe3}
  \hat{\theta}_{\modiff{x_c}|y,x_1,\cdots,x_n} &=&
  \frac{ n_{\modiff{x_c}|y,x_1,\cdots,x_n} + \hat{\phi}_{\modiff{x_c}|y,x_1,\cdots,x_{n-1}}\alpha_{y,x_1,\cdots,x_n}  }
       {n_{\cdot|y,x_1,\cdots,x_n} + \alpha_{y,x_1,\cdots,x_n}}
\end{eqnarray}

\subsection{Gibbs sampling}
Note, in Equation~\ref{eq-dm}, the counts $n_*$ are derived
quantities (summed from their child pseudo-counts)
and all pseudo-counts $t_*$ are latent variables that are sampled
using a Gibbs algorithm.
Moreover, the parameters $\theta_{\modiff{x_c}|y,x_1,\cdots,x_i}$ and
$\phi_{\modiff{x_c}|y,x_1,\cdots,x_i}$
are estimated recursively from $\phi_{\modiff{x_c}|y,x_1,\cdots,x_{i-1}}$
and the corresponding counts $n_{\modiff{x_c}|y,x_1,\cdots,x_i}$
using the standard CRP parameter estimation
of Equations~\ref{eq-crpe1}-\ref{eq-crpe3}.
Gibbs sampling of the pseudo-counts $t_*$
and the concentration parameters $\alpha_{*}$
is done and the estimation of $\theta_{\modiff{x_c}|y,x_1,\cdots,x_i}$ 
is made periodically to obtain an MCMC estimate for it.
This section then discusses how the Gibbs sampling of
these are done.

\subsubsection{Sampling pseudo-counts $t_*$}\label{subsubsec:pseudo-counts}
We use a direct strategy for sampling the $t_*$, sweeping
through the tree  sampling each pseudo-count individually
using a formula derived from Equation~\ref{eq-dm}:
\begin{eqnarray*}
\label{eq-dmt}
\lefteqn{\P(t_{\modiff{x_c}|y,x_1,\cdots,x_i}|\data,n_*,t_*^{-\modiff{x_c}|y,x_1,\cdots,x_i},\phi_{X},\alpha)
  ~\propto~~~~~~~}
  &&\\
\nonumber &&
\frac{\alpha_{y,x_1,\cdots,x_{i}}^{t_{\modiff{x_c}|y,x_1,\cdots,x_{i}}}}
      {\alpha_{y,x_1,\cdots,x_{i-1}}^{(n_{\cdot|y,x_1,\cdots,x_{i-1}})}}
   S^{n_{\modiff{x_c}|y,x_1,\cdots,x_{i-1}}}_{t_{\modiff{x_c}|y,x_1,\cdots,x_{i-1}}}  S^{n_{\modiff{x_c}|y,x_1,\cdots,x_{i}}}_{t_{\modiff{x_c}|y,x_1,\cdots,x_{i}}} ~,
\end{eqnarray*}
where $t_*^{-\modiff{x_c}|y,x_1,\cdots,x_i}$ represents
$t_* - \{ t_{\modiff{x_c}|y,x_1,\cdots,x_i}\}$.
\modiff{Note that $t_{\modiff{x_c}|y,x_1,\cdots,x_i}$ exists
implicitly in the two sums $n_{\cdot|y,x_1,\cdots,x_{i-1}}$ and $n_{\modiff{x_c}|y,x_1,\cdots,x_{i-1}}$ due to Equation~\ref{eq-crpstat}.}
This sweep is made efficient because computing the Stirling numbers is
a table lookup, and the Stirling numbers are
shared among the different trees, so they are only calculated once for all nodes of the BNC.


The base case, $i=0$ is different because
the root parameter vector $\phi_\modiff{X_c}$
is marginalised using the Dirichlet integral:
\[
\P(t_{\modiff{x_c}|y}|\data,n,t^{-\modiff{x_c}|y},\alpha) \propto
\frac{\Gamma\left(n_{\modiff{x_c}|y}+\alpha_0/|\modiff{X_c}|\right)}{\Gamma\left(n_{\cdot|y}+\alpha_0\right)}
\alpha_{y}^{t_{\modiff{x_c}|y}} S^{n_{\modiff{x_c}|y}}_{t_{\modiff{x_c}|y}} ~.
\]
These two sampling formula, as they stand, are also inefficient
because $t_{\modiff{x_c}|y,x_1,\cdots,x_i}$ ranges over
$1,\cdots,n_{\modiff{x_c}|y,x_1,\cdots,x_i}$ when $n_{\modiff{x_c}|y,x_1,\cdots,x_i}>0$.

From DP theory, we know that the pseudo-counts
$t_{\modiff{x_c}|y,x_1,\cdots,x_i}$ have a standard deviation given by
$O(\log^{1/2}n_{\modiff{x_c}|y,x_1,\cdots,x_i})$, which is very small,
thus in practice the full range is almost certainly never used.
Moreover, note the mean of $t_{\modiff{x_c}|y,x_1,\cdots,x_i}$ changes
with the concentration parameter, so in effect the
sampler is coupled and large moves in the ``search'' may not be effective.
As a safe and efficient option,
we only sample the pseudo-counts within
a window of $\pm 10$ of their current value.
We have tested this empirically, and due to the standard
deviations, it is safer as the Monte Carlo sampling converges
and smaller moves are typical.

Moreover, to initialise pseudo-counts in the Gibbs sampler, we use the
expected value of the pseudo-count for a HDP given the current
count and the relevant concentrations:
\begin{equation}\label{eq:init-t}
  t \leftarrow \left\{
    \begin{array}{ll}
      n &\text{ if }n\leqslant 1\\
      \max(1,\left \lfloor{\alpha \left( \psi_0(\alpha+n)-\psi_0(\alpha)\right)}\right \rfloor&\text{ if }n > 1\\
    \end{array}
  \right.
\end{equation}
This requires sweeping up the tree from the data at the leaves; \modiff{$\psi_0$ represents the digamma function: $\psi_0(x)=\frac{\Gamma '(x)}{\Gamma (x)}$.}


\subsubsection{Sampling and tying concentrations $\alpha_*$}
\modiff{No proper mention has been made yet of how the
concentration parameters $\alpha_*$ are sampled.
The concentration parameters influence how similar the child probability will be to
the parent probability.
We know this because Dirichlet theory tells us,
looking at the model in Section~\ref{ssec:model},
\[
\mbox{Variance}
(\theta_{\modiff{X_c}|y,x_1,\cdots,x_n})
\approx
\frac{1}{\alpha_{y,x_1,\cdots,x_{n}}}
\phi_{\modiff{X_c}|y,x_1,\cdots,x_{n-1}}(1-\phi_{\modiff{X_c}|y,x_1,\cdots,x_{n-1}})
\]
Since we cannot be sure of how large this will be, we also sample concentration. Experience with other models using HDPs alerts us that significant improvements should be possible by judicious sampling of the concentration parameters \cite{Buntine:2014}.}

\modiff{Note we expect
the variance to get smaller as we go down the tree,
so the concentration should be larger further down the tree.}

\label{subsubsec:sampling concentration}
\emph{Tying:} \modiff{Since the number of parameters $\alpha_*$ is equal to the number
of nodes in the tree, there are possibly too many to sample. So rather than using a separate concentration parameter
$\alpha_{\modiff{X_c}|y,x_1,\cdots,x_i}$ for every node, we tie some, which means that we make their values equal for some different nodes.}
\modif{Figures~\ref{fig:kdb1-HDP}(a) and Figure~\ref{fig:kdb1-HDP}(b) represent two different tying strategies of concentration parameters. }\modiff{The first one corresponds to tying the concentrations for all nodes that share a parent node: there will thus be a concentration parameter for all nodes in the tree but the lowest one. The second one has only one concentration parameter for each level of the tree. }\modif{Tying is only done within one context-tree, i.e. the parameters are inferred completely independently for each conditional probability distribution $\theta_{\modiff{X_i} | \Pi_i(\obj)}$.} Experiments on the tying of these hyperparameters
are presented in Section~\ref{subsec:exp-tying-iterations}.

\modiff{Note that the sampling described below iterates over all the tied nodes (see $j$); so different tying strategies only affect the nodes that the sampler runs over. }

\emph{Sampling:} \modiff{We use an augmentation
detailed in Section~4.3 of \cite{LimIJAR16}.
This introduces a new latent variable for each node},
and then a gamma sample can be taken for the
tied variable after summing the statistics across the tied nodes.
The general form of the likelihood for a concentration, $\alpha$,
from Equation~\ref{eq-dm} is $\prod_{j} \frac{\alpha^{t_j}}{\alpha^{(n_j)}}$
where $j$ runs over the tied nodes and  $(n_j,t_j)$ are the
corresponding counts at the nodes.
To sample $\alpha$ we need to augment the denominator terms
$\alpha^{(n_j)}$
because they have no match to a known distribution.
\modiff{This is done by adding a new term on both sides
$\P(q|\alpha,n)$ which introduces $q_j|\alpha \sim \mbox{Beta}(\alpha,n_j)$,
  then the joint posterior is derived as follows
  \begin{eqnarray*}
    \P(\alpha|\data,n,t)\P(q|\alpha,n)&\propto&\P(\alpha)\,
    \left(\prod_{j} \frac{\alpha^{t_j}}{\alpha^{(n_j)}}\right)
    \P(q|\alpha,n)\\
    \P(\alpha,q|\data,n,t)&\propto&
  \P(\alpha)\,
  \prod_{j}
  \frac{\alpha^{t_j}}{\alpha^{(n_j)}}
  \prod_j q_j^{\alpha-1}(1-q_j)^{n_j}\frac{\Gamma(\alpha+n_j)}{\Gamma(\alpha)\Gamma(n_j)}\\
  &\propto&
 \P(\alpha) \, \prod_{j} \alpha^{t_j}q_j^{\alpha-1}(1-q_j)^{n_j} ~.
  \end{eqnarray*}
  Looking closely at this, one can see $\alpha$ in the
  augmented distribution has a gamma likelihood.
  Thus, using a gamma prior
  $\alpha ~\sim \mbox{Gamma}(\nu_0,\mu_0)$ makes everything work simply.}  
The derived sampling algorithm for $\alpha$ is as following:
\begin{enumerate}
\item
  sample $q_j\sim \mbox{Beta}(\alpha,n_j)$ for all $j$, then
\item
  sample $\alpha \sim  \mbox{Gamma}\left(\nu_0+\sum_j t_j,\, \mu_0+\sum_j \log 1/q_j\right)$.
\end{enumerate}
Note for our experiments we use an empirical Bayesian approach,
so $\nu_0=\mu_0=0$, and leave the issue of selecting an appropriate prior
as further research.

\subsection{\modif{Algorithmic description}}
\modif{We present here a high-level description of our sampler and associated HDP-estimates in Algorithms~\ref{alg:init} to \ref{alg:sample-concentration}.}

\modif{Algorithm~\ref{alg:init} is the main algorithm: it takes as an input a dataset and returns a context tree containing our HDP estimate. It starts by creating the tree based on the dataset, i.e. creating the branches for the different cases present in the dataset, as well as storing the count statistics at the leaves. The tree is a typical hierarchical structure with a root node; nodes contain the count statistics $t_\star$ and $n_\star$, a link to its concentration $\alpha$ and a link to a table of children (one child per value of the branching variable at that node). It then calls the initialisation of the pseudo-counts $t_\star$ in the tree, and creates an array of concentration parameters that are tied at each level.
It then proceeds with the sampling process. For each iteration of the sampling process, we first sample the $t_\star$ from the leaves up to the root, then we sample the concentration parameters (one per level except for the root node, which is not sampled).
Finally, after the burn-in period has passed, we record and average the probability estimates in the tree at the current iteration. When the sampling process is terminated, these averaged estimates (stored in the tree)
constitute our HDP estimates; they can be accessed by querying the tree. For brevity, we do not describe the following simple functions:
  \begin{itemize}
  \item \texttt{getNodesAtDepth}: returning all nodes at a given depth of the tree
  \item \texttt{initTreeWithDataset}: creating the branches of the tree down to the leaves for which data exists
  \item \texttt{createConcentrationArray}: creating an array of concentration objects of given size
  \item \texttt{recordProbabilityRecursively}: averaging the estimates for all nodes in the tree
  \end{itemize}}
\begin{algorithm2e}\modif{
    \DontPrintSemicolon
\KwIn{$\data$: the dataset}
\KwIn{nIters: number of iterations to run the sampler for}
\KwIn{nBurnIn: number of burn-in iterations before starting to average out the $\theta$s}
\caption{\label{alg:init}EstimateProbHDB(data, nIters, nBurnIn)}
tree $\leftarrow$ initTreeWithDataset($\data$) \tcp*{create tree with avail. data}
initParametersRecursively(tree.root)\tcp*{Algorithm~\ref{alg:init-node}}
\tcp{table of concentrations, one per level (\texttt{Level} tying)}
cTab $\leftarrow$ createConcentrationArray(tree.depth)\;
\For{\emph{depth} $\gets 1$ \KwTo \emph{tree.depth}}{
  \ForEach{\emph{node} $\in$ \emph{tree.getNodesAtDepth(depth)}}{
    node.$\alpha\gets$cTab[depth]\;
  }
}
\For(\tcp*[f]{Gibbs sampler}){\emph{iter} $\leftarrow 1$ \KwTo \emph{nIters}}{
  \tcp{sampling parameters for all nodes bottom-up}
  \For{\emph{depth} $\leftarrow$ \emph{tree.depth} \KwTo $1$}{
    \ForEach{\emph{node} $\in$ \emph{tree.getNodesAtDepth(depth)}}{
      sampleNode(node,10,cTab[depth]) \tcp*{Algorithm~\ref{alg:sample-node}}
    }
  }
  \For(\tcp*[f]{sampling concentrations}){$\text{level}\leftarrow 2$ \KwTo \emph{tree.depth}}{
    sampleConcentration($\alpha$,tree.getNodesAtDepth(level))\tcp*{Algorithm~\ref{alg:sample-concentration}}
  }
  \If{\emph{iter} $>$ \emph{nBurnIn}}{
    recordProbabilityRecursively(tree.root)
  }
}
\Return tree\;
}\end{algorithm2e}

\modif{Algorithm~\ref{alg:init-node} describes the initialisation process of the tree's statistics, which is performed bottom-up. Starting from the leaves, we propagate the pseudo-count $t_\star$, which constitutes the $n_\star$ statistics of parent nodes (lines 1--9). Initialisation of the pseudo-counts $t_\star$ is done following Equation~\ref{eq:init-t}. }
\begin{algorithm2e}\modif{
     \DontPrintSemicolon
\KwIn{node: node of which we want to initialise the parameters}
\caption{\label{alg:init-node}initParametersRecursively($node$)}
\If(\tcp*[f]{init. children and collect stats}){\emph{node is \textbf{not} a leaf}}{
  \ForEach{\emph{child} $\in$ \emph{node.children}}{
    initParametersRecursively(child)\;
    \For{$k \gets 1$ \KwTo $\modiff{|X_c|}$}{
      node$.n[k] \gets $node$.n[k] + $child$.t[k]$\;
      node$.n \gets $node$.n + $node$.n[k]$ \tcp*{marginal}
    }
  }
}
\eIf{\emph{node is root}}{
  $\forall k, \text{node}.t[k]\gets \min(1,\text{node}.n[k])$
}{
  \For{$k \gets 1$ \KwTo $\modiff{|X_c|}$}{
    \eIf{\emph{node}$.n[k]\leqslant 1$}{
      $\text{node}.t[k]\gets \text{node}.n[k]$
    }{
      $\text{node}.t[k]\gets \max(1,\left \lfloor{\text{node}.\alpha \left( \psi_0(\text{node}.\alpha+\text{node}.n)-\psi_0(\text{node}.\alpha)\right)}\right \rfloor)$
    }
  }
}
$\text{node}.t \gets \sum_k \text{node}.t[k]$ \tcp*{marginal}
}\end{algorithm2e}

\modif{Algorithm~\ref{alg:sample-node} describes the sampling of the pseudo-counts $t_\star$ associated with a node, i.e. the data that should be propagated up to the parent node. Sampling happens if and only if the node is not the root node, and the $n_\star$ count statistics are strictly greater than 1.\footnote{\modif{If $n_\star=0$, then no data has been propagated from the children, and hence no data can be propagated up to the parent. If $n_\star=1$, then that datapoint has to be propagated to the parent and hence needs no sampling.}} The pseudo count is then sampled using the window described in Section~\ref{subsubsec:pseudo-counts}; values either outside this window, or impossible given the pseudo-count at the parent get assigned a 0 probability of being sampled (see Algorithm~\ref{alg:change-tk-get-progbability}). Valid values within the window are sampled following the Equations presented in Section~\ref{subsubsec:pseudo-counts}. 
}
\begin{algorithm2e}\modif{
     \DontPrintSemicolon
\KwIn{node: node of which we want to sample the parameters}
\KwIn{$w$: window for sampling}
\KwIn{$\alpha$: concentration to assign to node}
\caption{\label{alg:sample-node}sampleNode($node,w,\alpha$)}
\eIf{\emph{node is root}}{
  $\forall k, \text{node}.t[k]\gets \min(1,\text{node}.n[k])$ \tcp*{no sampling}
}{
  node.$\alpha\gets\alpha$\tcp*{assign concentration to node}
  \For{$k \gets 1 \cdots \modiff{|X_c|}$}{
    \eIf{\emph{node}$.n[k]\leqslant 1$}{
      $\text{node}.t[k]\gets \text{node}.n[k]$ \tcp*{value fixed}
    }{
      $\text{minTk} \gets \max\left(1, \text{node}.t[k]-w  \right)$\;
      $\text{maxTk} \gets \min\left( \text{node}.t[k]+w,\text{node}.n[k] \right)$\;
      \tcp{Constructing a vector to sample node$.t[k]$ from}
      $\vec{v}\gets \vec{0}$ \tcp*{length is $(\text{node}.n[k]+1)$}
      \For{$t\gets$ \emph{minTk} $\cdots$ \emph{maxTk}}{
        $\vec{v}_t\gets \text{changeTkAndGetProbability}(\text{node},k,t)$\tcp*{Algorithm~\ref{alg:change-tk-get-progbability}}
      }
      $\forall t,\vec{v}_t\gets \frac{\vec{v}_t}{\sum_t \vec{v}_t}$\tcp*{Normalize vector}
      $t \sim \mbox{multinomial}\left( \vec{v} \right)$\;
      $\text{changeTkAndGetProbability}(\text{node},k,t)$\tcp*{Algorithm~\ref{alg:change-tk-get-progbability}}
    }
  }
}
}\end{algorithm2e}

\modif{Algorithm~\ref{alg:change-tk-get-progbability} both changes the value of a pseudo-count $t_\star$ at a node and returns its probability. As described above, it starts by checking that the new value for the pseudo-count is valid (else does not do the change and return probability 0). It then updates the pseudo-count for that node, and the count statistic $n_\star$ at the parent. It finally returns the probability as described in Section~\ref{subsubsec:pseudo-counts}.
}
\begin{algorithm2e}\modif{
     \DontPrintSemicolon
\KwIn{node: node of which we want to sample the parameters}
\KwIn{\modiff{$k$: index of the value we want to change in $t$}}
\KwIn{\modiff{newValue: value to replace $t_k$ by, if possible}}
\caption{\label{alg:change-tk-get-progbability}changeTkAndGetProbability(node,$k$,newValue)}
$\text{inc} \gets \text{newValue} - \text{node}.t[k]$\;
\If(\tcp*[f]{check if valid for parent}){\emph{inc}$ < 0$}{
  \If{\emph{node is not root \textbf{and}} $($\emph{node.parent}$.n[k]+$\emph{inc}$)<~$\emph{node.parent}$.t[k]$}{\Return 0\;}
}
$\text{node}.t[k]\gets \text{node}.t[k] + \text{inc}$\;
$\text{node}.t\gets \text{node}.t + \text{inc}$ \tcp*{marginal}
\If(\tcp*[f]{update statistics at the parent}){\emph{node is not root}}{
  $\text{node.parent}.n[k]\gets \text{node.parent}.n[k] + \text{inc}$\;
  $\text{node.parent}.n\gets \text{node.parent}.n + \text{inc}$ \tcp*{marginal}
}
\Return $\frac{\text{node}.\alpha^{\text{node}.t[k]}\cdot S^{\text{node.parent}.n[k]}_{\text{node.parent}.t[k]}\cdot S^{\text{node}.n[k]}_{\text{node}.t[k]}}{rising\_factorial(\text{node.parent}.\alpha,\text{node.parent}.n[k])}$
}\end{algorithm2e}

\modif{
  Finally, Algorithm~\ref{alg:sample-concentration} describes a simple sampling of the concentration parameters in the tree, assuming that tying is done using the \texttt{Level} strategy. As described in Section~\ref{subsubsec:sampling concentration}, tying requires to iterate through the $t_\star$ and $n_\star$ of the `tied' nodes. For all the `tied' nodes, it thus performs a change of variable to $q$ and then samples the new concentration. Other tying strategies are given in the source-code function \texttt{Concentration.java:sample()} \modiff{(see beginning of Section 6.1 for link to source code)}.
}
\begin{algorithm2e}\modif{
     \DontPrintSemicolon
\KwIn{$\alpha$: concentration to sample}
\KwIn{nodes: nodes sharing this concentration parameter (tying)}
\caption{\label{alg:sample-concentration}sampleConcentration($\alpha, \text{nodes}$)}
rate $\gets 0$\;
\ForEach{\emph{node} $\in$ \emph{nodes}}{
  $q\sim \mbox{Beta}(\alpha,\text{node}.n)$\tcp*{change of variable, sample $q$}
  rate $\gets$ rate $-\log(q)$\;
}
$\alpha\sim  \mbox{Gamma}\left(\sum_{n\in nodes}n.t,\, \text{rate}\right)$\tcp*{sample $\alpha$}
\ForEach(\tcp*[f]{assign new $\alpha$ to nodes}){\emph{node} $\in$ \emph{nodes}}{node.$\alpha\gets\alpha$}
}\end{algorithm2e}

\subsection{\modif{Worked example}}
\label{subsec:Worked example}
\modif{
  We have now fully described our HDP-based estimates.
  In this section, we draw all the theory together and show how our method applies to two simple datasets, highlighted in Table~\ref{tab:worked-examples}. 
  Both datasets have two binary variables $X_1$ and $Y$, and a simple na\"{i}ve Bayes structure, \ie{}we focus on the estimation of $\P(X_1|Y)$. 
  Although this simple structure does not give full justice to our estimates for deeper hierarchies, we feel that such an example helps understanding the different components of our method.
}
\begin{table}[H]\centering
  \modiff{
    \begin{tabular}{lccccc}
      Dataset&Value&Frequency&\multicolumn{3}{c}{$\hat{p}(X_1|Y)$}\\\cline{4-6}
      &for $Y$&$n_{X_1|y}$&MLE&$m$-estimate ($m=1$) &\textbf{HDP}\\\hline
      1&$0$&$[2,0]$ & $[1.00,0.00]$ & $[0.83,0.17]$ & $[0.89,0.11]$\\
      &$1$& $[20,5]$ & $[0.80,0.20]$ & $[0.79,0.21]$ & $[0.79,0.20]$\\\hline
      2&$0$&$[2,0]$ & $[1.00,0.00]$ & $[0.83,0.17]$ & $[0.86,0.14]$\\
      &$1$&$[4,9]$ & $[0.31,0.69]$ & $[0.32,0.68]$ & $[0.34,0.66]$
    \end{tabular}
    \caption{\label{tab:worked-examples}Example datasets with associated estimates}
  }
\end{table}

\modif{Our aim is to highlight how information is shared between $\P(X_1|Y=0)$ and $\P(X_1|Y=1)$ through the marginal (mean) probability $\P(X_1)$. Let us describe the two datasets given in Table~\ref{tab:worked-examples}: Dataset~\#1 has $\P(X_1|Y=0)\approx \P(X_1|Y=1)$ --~but with only little data available to estimate $\P(X_1|Y=0)$~-- while Dataset~\#2 has $\P(X_1|Y=0) \not\approx \P(X_1|Y=1)$.}

\modif{Let us start by the analysis of the cases with $Y=0$ compared for the two datasets, cases for which the data available is identical. The first thing to observe is that, as the frequency is the same for both datasets for the cases with $Y=0$, so are the MLEs and $m$-estimates\footnote{More information about $m$-estimates is given in Section~\ref{subsec:exp-settings}.}, respectively. MLEs and $m$-estimates are agnostic of the marginal; $m$-estimates only pull the estimates toward a uniform prior. Second, we can observe that our HDP estimates for Dataset~\#1 are closer to the MLEs than to the $m$-estimates. This is because the data available for $Y=1$ `corroborates' the fact that $\P(X_1=0|Y)$ is much greater than $\P(X_1=1|Y)$. For Dataset~\#2 where the two cases for $Y$ differ, we can see that our estimate for $Y=0$ is closer to the $m$-estimate than it was for Dataset~\#1 although the frequencies are the same; this is because now the data available for $Y=1$ does not support the hypothesis that the marginal $\P(X_1)$ is helpful to estimate $\P(X_1|Y=0)$ while having little data available. Finally, we can see that our HDP estimate for $\P(X_1|Y=1)$ in Dataset~\#2 goes even further than the $m$-estimate and pulls the estimate even closer to a uniform probability. This is again here because of the data for $Y=0$.
}

    

\section{Related work}
\label{sec:Related work}
Extensive discussions of methods for DP and PYP hierarchies
are presented by \citet{GasthausTeh10,LimIJAR16}.
Standard Chinese restaurant process (CRP) samplers \cite{Teh2006}
use dynamic memory so are computationally demanding,
and not being collapsed also makes them considerably slower.
\citet{LimIJAR16} deal with the case where the counts at the
leaves of the tree are latent, and thus are not applicable to our context.
The direct samplers of \citet{du2010segmented},
which are also \emph{collapsed} CRP samplers,
are more efficient than CRP samplers and those of \citet{LimIJAR16} in the current context.
\citet{GasthausTeh10} dealt with a PYP where the discount
parameters change frequently so direct samplers were inefficient
because the cache of Stirling numbers needed constant recomputation.
On-the-fly samplers have also been developed by \citet{Shareghi2017PYP} for PYP hierarchies, making it possible to use PYP for deep trees and large dataset sizes. 
This however does not change the issue of constant recomputation of Stirling numbers, which is why initialisations based on modified Kneser-Ney have been developed by \citet{Shareghi2016PYP}. 

The use of DP and PYP hierarchies for regression and clustering --~as opposed to classification in our case~-- has been studied by \citet{nguyen2015Bayesian,huynh2016scalable}, respectively. 

Related work for BNCs was discussed in \ref{sec:structure-learning}. There are other methods for improving BNCs. A simple back-off strategy, backing off to the root, is proposed by \citet{friedman:bnc}. Moreover, for some simple classes of networks, such as TAN, a disciminative generalisation of logistic regression can be used because the optimisation surface is convex \cite{Roos2005,zaidi2017efficient}.
Neither techniques are applicable to the more complex BNCs we consider.

Bayesian model averaging methods are common for Bayesian network learning \cite{friedman2003being}. 
Average n-dependence estimators -- AnDE \cite{WebbBoughtonWang05,WebbEtAl12},
another ensemble method, is competitive for smaller data sets but
cannot compete against SkDB for larger data sets \cite{Martinez2016}.

Either way, these invariably use the same Laplacian prior
as the m-estimates reported here in Section~\ref{sec:Experiments}.


\section{Experiments}
\label{sec:Experiments}
The aim of this section is to assess our HDP-based estimates for Bayesian network classifiers (BNCs). 
In Section~\ref{subsec:exp-settings}, we give the general settings that are necessary to understand and reproduce our experiments. Then, in Section~\ref{subsec:exp-tying-iterations}, we start by studying how to parameterize our method: i.e.\ by studying the influence of number of iterations and the tying strategy used. In Section~\ref{subsec:exp-BN-vs-BN}, we demonstrate the superiority of our estimates over the state of the art across 8 different BNC structures. Finally, having obtained significant improvements over the state-of-the-art, we then turn to comparing the best-performing configuration (TAN and SkDB with HDP estimates) with random forest (RF) in Section~\ref{subsec:exp-BN-vs-RF}. We show that our estimate allows even models as simple as TAN to significantly outperform RF (with statistical significance), while standard approaches to parameter estimation are beaten by RF. We conclude the experiments with a demonstration of our system's out-of-core capability and show results obtained on the Splice dataset with 50 million training examples, a quantity that RF cannot handle on most machines. 

\subsection{Experimental design and setting}
\label{subsec:exp-settings}
\emph{Design:} All experiments are carried out on a total of $68$ datasets from the UCI archive \cite{UCI-ML}; $38$ datasets with less than $1000$ instances, $23$ datasets with instances between $1000$ and $10000$, and $7$ datasets with more than $10000$ instances. 
The list and description of the datasets is given in Table~\ref{UCIDatasets} at the end of this paper. 
For all methods, numeric attributes are discretized by using the minimum description length (MDL) discretization method~\cite{Fayyad1992}. A missing value is treated as a separate attribute value and taken into account exactly like other values.
Each algorithm is tested on each dataset using $2$-fold cross validation
repeated $5$ times.
We assess the results by reporting 0-1 Loss and RMSE, and report Win-Draw-Loss (W-D-L) results when comparing the 0-1 Loss and RMSE of two models. 
A two-tail binomial sign test is used to determine the significance of the results, using $p \leq 0.05$.

Note the RMSE is related to the Brier score, which is a proper scoring
rule for classifiers and thus generally preferable to error, especially
in the context of unequally occurring classes or unequal costs.
It measures how well calibrated the probability estimates are.
We use it because we suspected that our methods could improve
probability estimates but not necessarily errors.

\emph{Software: }To ensure reproducibility of our work and allow other researchers to easily build on our research, we have made our source code for HDP parameter estimation available on \href{https://github.com/fpetitjean/HierarchicalDirichletProcessEstimation}{Github}. 

\emph{Compared methods: }
We assess our estimates for 8 BNC structures with growing complexity. Our BNC structures are: na\"{i}ve Bayes (NB), tree-augmented na\"{i}ve Bayes (TAN) \cite{friedman:bnc}, k-dependence Bayesian network (kDB) \cite{sahami:lldbc} with $k=1\text{ to }5$ and selective kDB (SkDB) \cite{Martinez2016} with maximum $k$ set to 5 also.\footnote{\modif{We do not consider higher values of $k$, because (1) for kDB we will see in Section~\ref{subsec:exp-BN-vs-BN} that the superiority of our HDP estimates is statistically significant further increases with $k$; (2) for SkDB, 95\% of the experiments see it choose a structure with $k<5$, differences with higher $k$ would thus be minimal.}} When comparing to random forest (RF), we use the Weka default parameterization, i.e.\ selecting $\log_2(n)+1$ attributes in each tree,\footnote{Selecting $\sqrt{n}$ attributes produces similar results and conclusion, so the results are left out of this paper for concision.} no minimum leaf size and using $100$ decision trees in this work.

For BNCs, we compare our HDP estimates to so-called m-estimates\footnote{Also known as Schurmann-Grassberger's Law when $m=1$, which is a particular case of Lidstone's law \cite{lidstone20,hardy} with $\lambda=\frac{1}{|X_i|}$, also based on a Dirichlet prior.}
\cite{mitchellbook} as follows:
 \begin{equation}
  \hat{p}(x_i|\Pi{(i)}) = \frac{counts(x_i,\Pi{(i)})+\frac{m}{|X_i|}}{counts(\Pi{(i)})+m}
 \end{equation}
 where $\Pi{(i)}$ are the parent-values of $X_i$. The value of $m$ is set by cross-validation on a holdout set of size $\min(N/10,5000)$ among with $m\in\{0,0.05,0.2,1,5,20\}$. 

Count statistics are stored in a prefix tree; for m-estimates, if zero counts are found, we back off as many levels in the tree as necessary to find at least one count. For instance, if $counts(x_4,x_0,x_3)$ is equal to zero, then $\hat{p}(x_4|x_0)$ is considered instead of $\hat{p}(x_4|x_0,x_3)$. Note that not using this strategy significantly degrades the performance of BNCs when using m-estimates (for our HDP estimates, the intermediate nodes $\phi$ are considered latent and thus inferred directly during sampling). 
 
\subsection{Tying and number of iterations}
\label{subsec:exp-tying-iterations}
Before proceeding with the comparison of our method to the state of the art, it is important to study two elements: (1) for how many iterations to run the sampler and (2) how to tie the concentration parameters. These two elements are directly related because the less tying, the more parameters to infer, which means that we expect to have to run the sampler for more iterations. 

\noindent
We consider three different tying strategies:
\begin{enumerate}
\item Same Parent (SP): children of each node share the same parameter -- illustrated in Figure~\ref{fig:kdb1-HDP}(a). 
\item Level (L): we use one parameter for each level of the tree -- illustrated in Figure~\ref{fig:kdb1-HDP}(b). 
\item Single (S): all parameters tied together.
\end{enumerate}

\begin{figure}
	\centering
    \includegraphics[width=.6\linewidth]{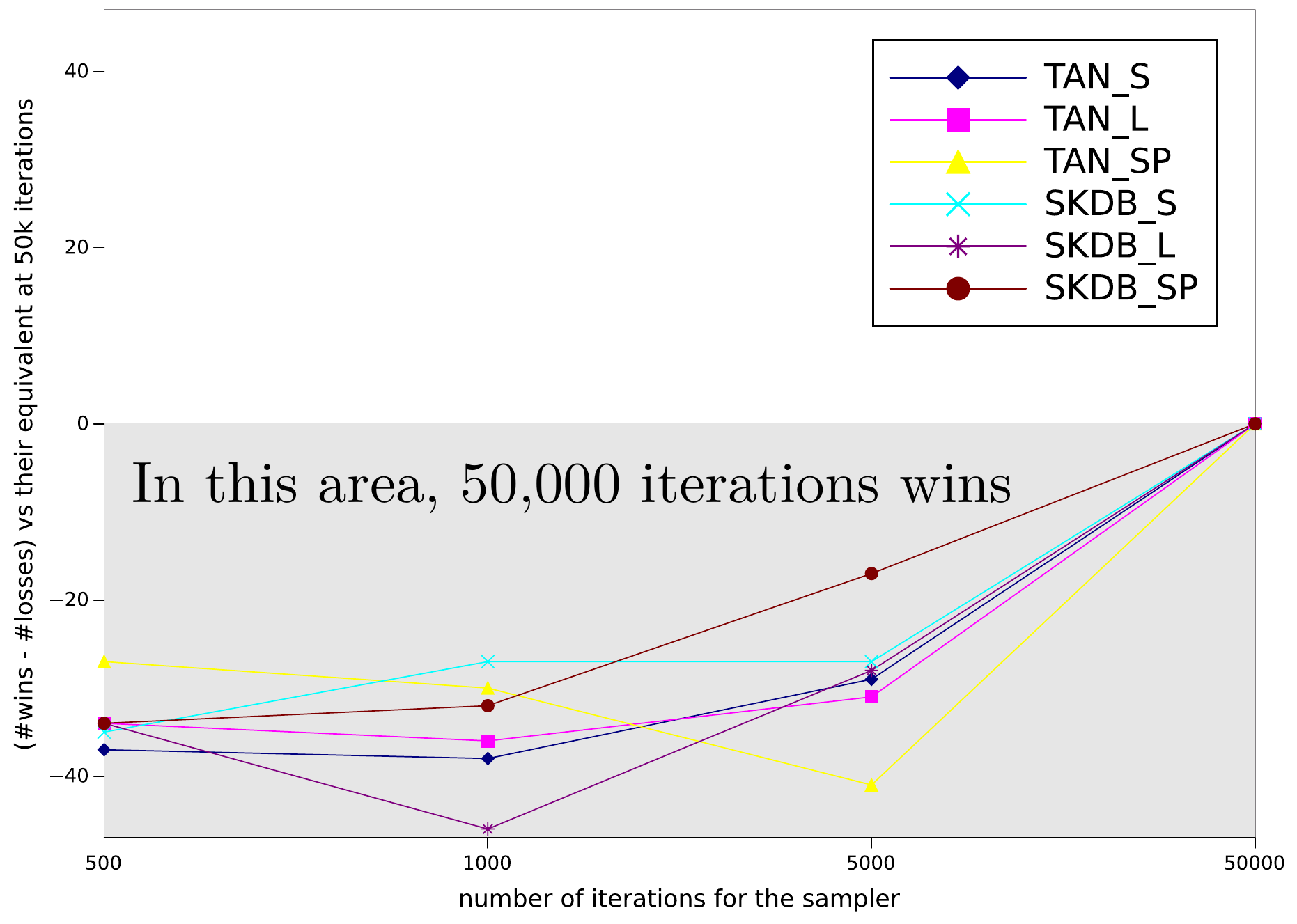}
    \caption{\label{fig:nIterations}Win/loss plot on RMSE for each combination of (flagship classifier) $\times$ (tying strategy). Comparison is for running each combination for $x$ iterations vs 50,000 and include \textbf{S}ingle, \textbf{L}evel and \textbf{S}ame\textbf{P}arent.}
\end{figure}

\emph{Number of iterations: }Asymptotically, the accuracy of the estimates improves as we increase the number of iterations. The question is how quickly they asymptote. We thus studied the performance of our two flagship classifiers --~TAN and SkDB~-- on all datasets as we increase the number of iterations from 500 to 50,000. For each combination of classifier$\times$tying strategy, we assess the win-loss profile for $x$ iterations versus 50,000. The resulting win-loss plot in Figure~\ref{fig:nIterations} shows that across all tying strategies and models, running our sampler for 50,000 iterations is significantly better than with fewer iterations. Even for models as simple as TAN with a Single concentration parameter, running the sampler for 5,000 iterations wins 13 times and loses 42 times as compared to running it for 50,000 iterations. Unless specified otherwise, we thus run the sampler for 50,000 iterations. We surmise that even more iterations could further improve accuracy but leave this for future research.

\begin{figure}
	\centering
    \includegraphics[width=.6\linewidth]{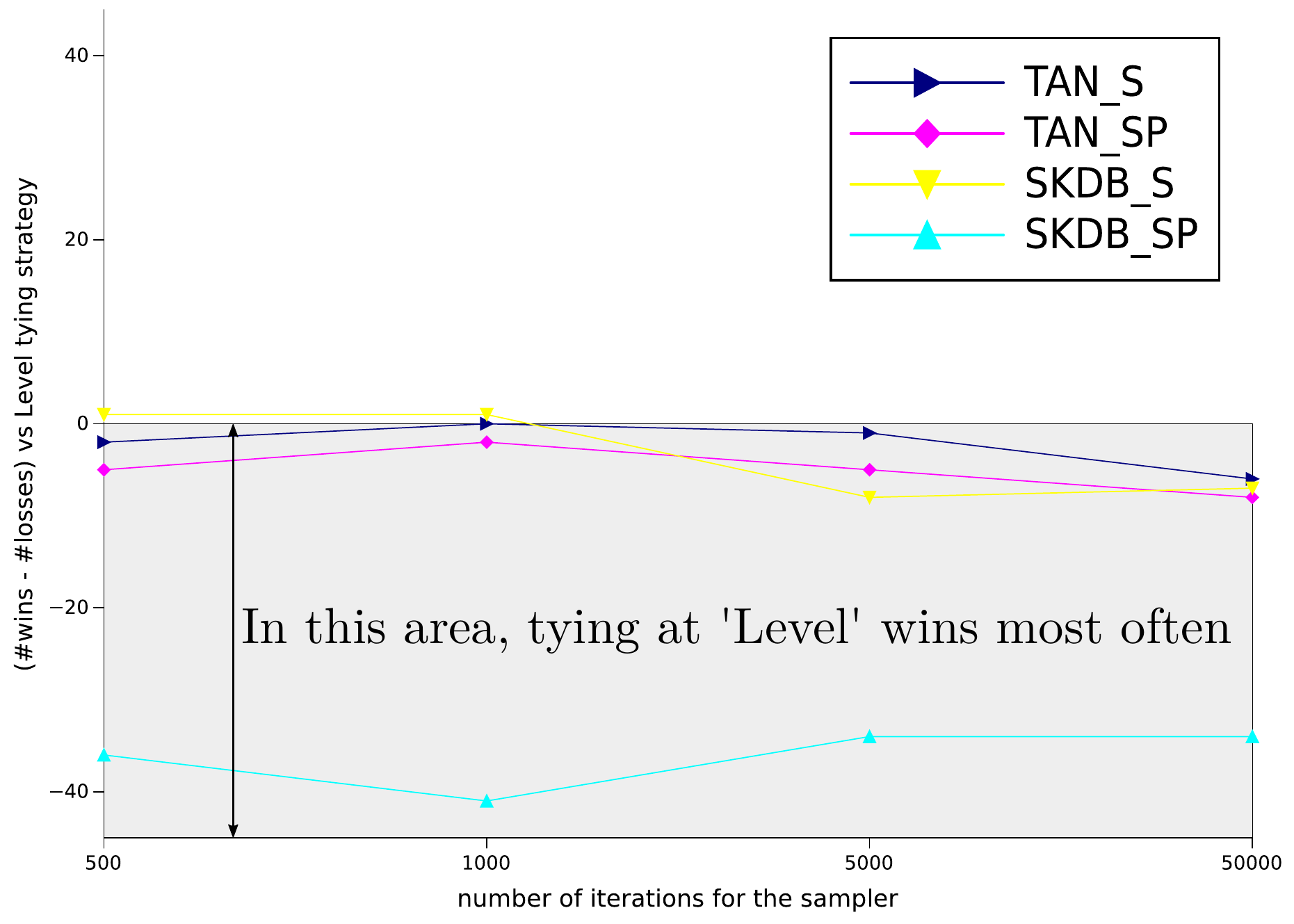}
    \caption{\label{fig:tying-vs-L}Win/loss plot of each combination of (flagship classifier) $\times$ (S or SP tying strategy) versus tying at level (L). }
\end{figure}

\emph{Tying strategy: }
Having seen that 50,000 iterations seems important regardless of the tying strategy, we here show that tying per Level seems to be the best default strategy. It is important to note that we do not intend to give a definitive answer valid for all domains here, but are simply giving a reasonable `default' parameterization. The Level strategy was illustrated for kDB-1 in Figure~\ref{fig:kdb1-HDP}(b).
To illustrate this we compare TAN and SkDB parameterized with the \emph{same parent} (SP) and \emph{single} (S) strategies versus using the \emph{level} (L) tying strategy across different numbers of iterations. Figure~\ref{fig:tying-vs-L} gives the win-loss plot. We see that L provides a uniformly good solution providing both the best results with 50,000 iterations but also providing solid performances as early as 500 iterations. It is worth noting that for TAN, the L and S strategies are very similar, only differing by one concentration parameter. The SP strategy seems to clearly underperform L, all the more when the complexity of the model increases, which makes sense given that the number of concentration parameters to estimate increases exponentially with the depth of the prefix tree, which is mostly controlled by the number of parents for each node $i$.
It is possible that for large amounts of data, the SP strategy would offer a better bias/variance tradeoff but such a study falls out of the scope of this paper. We thus use L as a tying strategy for the remainder of this paper. 
 
\subsection{HDP vs m-estimates for Bayes network classifiers}
\label{subsec:exp-BN-vs-BN}
So far, we have only assessed the relative performance of HDP estimates with different parameterizations. Having settled on  50,000 iterations and per Level tying, we now turn to the full comparison with the state-of-the-art in smoothing Bayesian network classifiers: using m-estimates with the value of $m$ cross-validated on a holdout set. We also remind the reader that, to provide the best competitor, we also added the back-off strategy described above, without which m-estimates cannot compete at all.

\begin{figure}
	\centering
    \subfigure[\label{fig:scatter-kDB-5}]{
    \includegraphics[width=.48\linewidth]{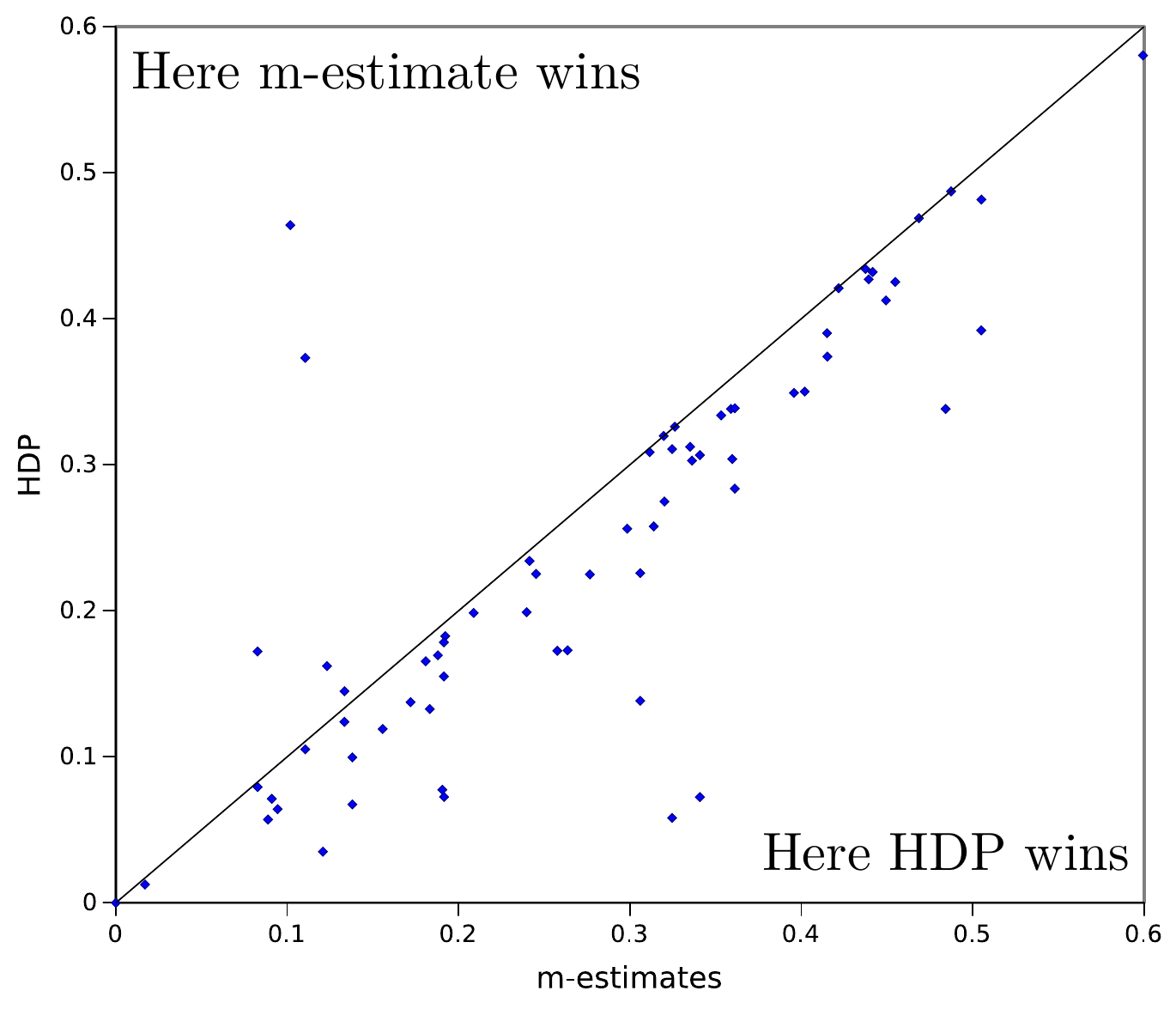}
    }
    \subfigure[\label{fig:kDB-5-overfitting}]{\includegraphics[width=.48\linewidth]{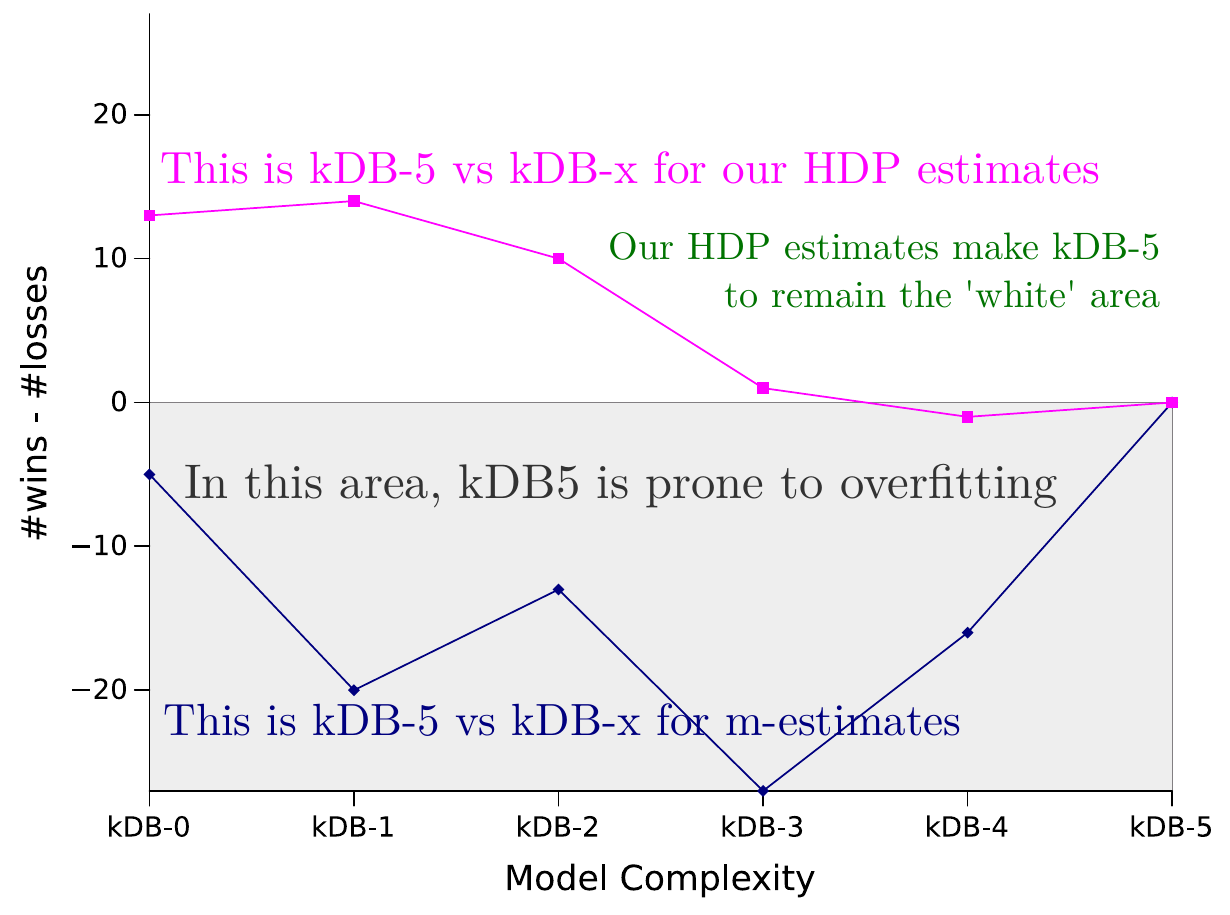}}
    \caption{(a) Scatter plot on RMSE for kDB-5 for HDP vs m-estimate. (b) Win/loss plot of kDB-5 vs kDB-$x$ for m-estimates vs our HDP ones. }
\end{figure}

We report in Table~\ref{tab:BN-vs-BN} the win-draw-loss of our HDP estimates versus m-estimates across 8 different BNCs from na\"{i}ve Bayes and TAN to kDB with $1\leqslant k\leqslant 5$ and SkDB. 
\begin{table}
	\centering
	\caption{\label{tab:BN-vs-BN}Win/Draw/Loss for 8 BNCs for our HDP estimate vs m-estimate. Stat. sig. ($p<0.05$) results are depicted in boldface. }
	\begin{tabular}{lll}
    	\hline \\[-5pt]
        Classifier&\multicolumn{2}{l}{Win--draw--loss for HDP vs m-estimate}\\[2pt]
		\cline{2-3} \\[-5pt]
		 &0/1-loss & RMSE \\[2pt]
		\hline \\[-5pt]
		\emph{Naive Bayes}  &\textbf{41--4--23}&40--0--28\\[2pt]
		\emph{TAN} & \textbf{45--4--19} & \textbf{52--1--15} \\[2pt]
		\emph{kDB-1} &\textbf{45--4--19}&\textbf{50--1--17}\\[2pt]
		\emph{kDB-2} &\textbf{54--2--12}&\textbf{54--0--14}\\[2pt]
		\emph{kDB-3} &\textbf{52--4--12}&\textbf{53--2--13}\\[2pt]
		\emph{kDB-4} &\textbf{56--4--\textcolor{white}{0}8}&\textbf{56--0--12}\\[2pt]
		\emph{kDB-5} &\textbf{60--4--\textcolor{white}{0}4}&\textbf{60--2--\textcolor{white}{0}6}\\[2pt]
		\emph{SkDB} & \textbf{45--4--19} & \textbf{54--0--14} \\[2pt]
		\hline
	\end{tabular}
\end{table}
It is clear from this table that our HDP estimates are far superior to m-estimates. It is even quite surprising to see our estimates outperform m-estimates with models as simple as Na\"{i}ve Bayes, where our hierarchy only has one single level. Moreover, as the model complexity increases (the maximum number of parents for each node), this difference increases. The scatter-plot for kDB-5 HDP vs m-estimate is given in Figure~\ref{fig:scatter-kDB-5} and shows again the same trend with HDP significantly outperforming m-estimate. As usual when dealing with a broad range of datasets, there are a few points for which HDP loses. Interestingly, the most important loss is for the \texttt{Cylinder-Bands} dataset, which contain only 540 samples, and thus for which we would have expected that smoothing would be important; detailed inspection of this dataset show that the 540 cases seem to be relatively similar to each other (in which case the cross-validation used for m-estimates help discover this). 

It is also interesting to study the capacity of HDP to prevent overfitting as compared to the m-estimate (with $m$ cross-validated). 
In Figure~\ref{fig:kDB-5-overfitting}, we report for m-estimates the win-loss plot for kDB-5 compared to kDBs with increasing complexity from 0 (kDB-0 is NB) to 4. Given that kDB-5 has generally lower bias than kDB $\forall k\leqslant 4$, we can typically attribute its losses to overfitting. Starting with the bottom line, which represents the behaviour of using m-estimates, we can see that kDB-5 generally loses to lower complexity kDBs. The maximum difference is with kDB-3 which seems to globally have a nice bias/variance tradeoff on this collection of datasets.

Conversely, we can see that HDP estimates (top-curve in Figure~\ref{fig:kDB-5-overfitting})  allows us to nicely control for overfitting.  What happens is that we make the most of the low-biased structure offered by kDB, while not being overly prone to overfitting. In some sense, our hierarchical process makes it possible to pull the probability estimates towards higher-level nodes for which we have more data, and this automatically depending on the dataset.  It seems that it makes it possible to be less strict about the structure and to be powerful at controlling for the variance. In fact, controlling for overfitting is what selective kDB (SkDB) tries to achieve; in our experiments, kDB5-HDP has a slight edge over SkDB5-HDP with a win-draw-loss of 33--5--30 on RMSE. Nevertheless, it remains that HDP largely outperforms m-estimates with a win-loss --~for SkDB~-- of 60 to 8. 

\begin{figure}
	\centering
\includegraphics[width=0.48\linewidth]{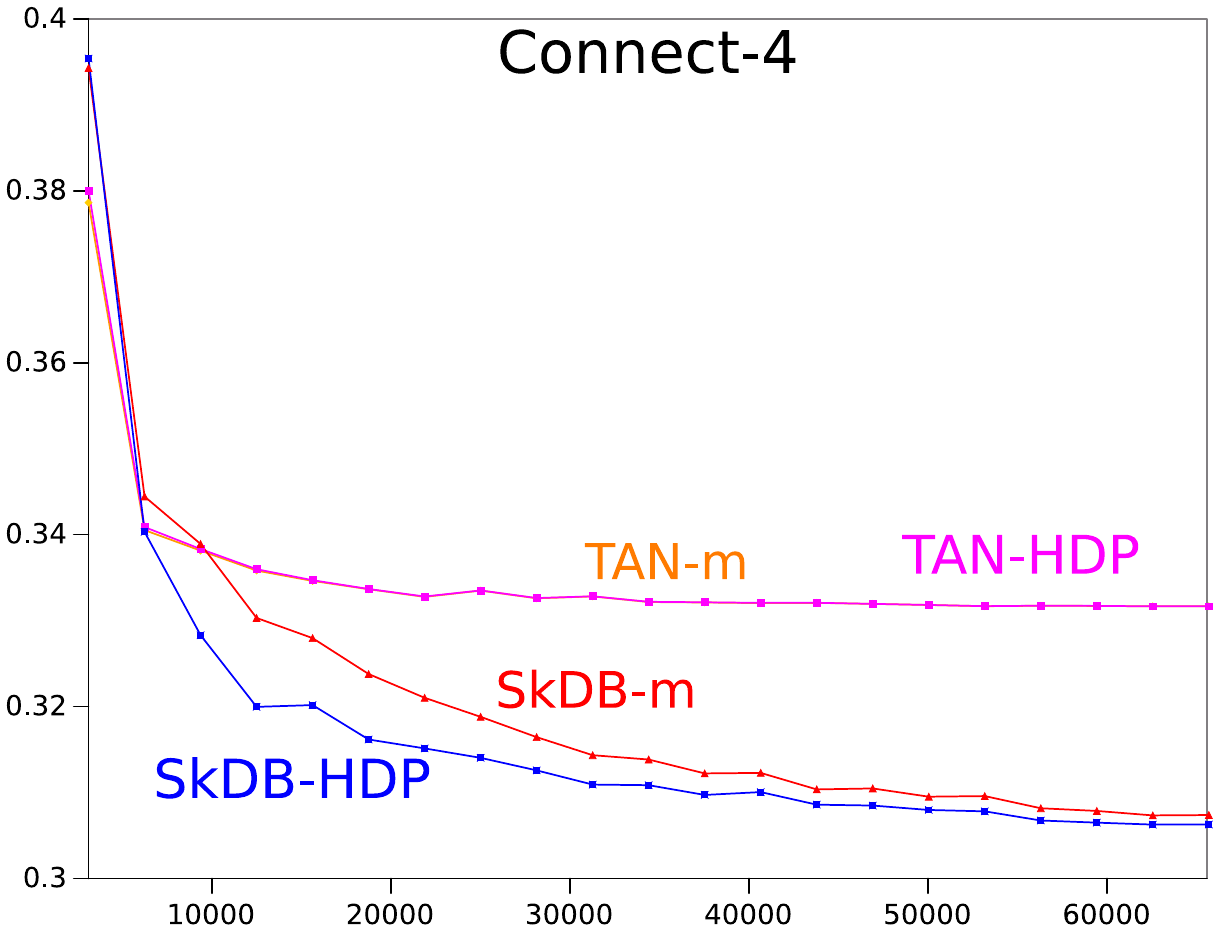}\includegraphics[width=0.48\linewidth]{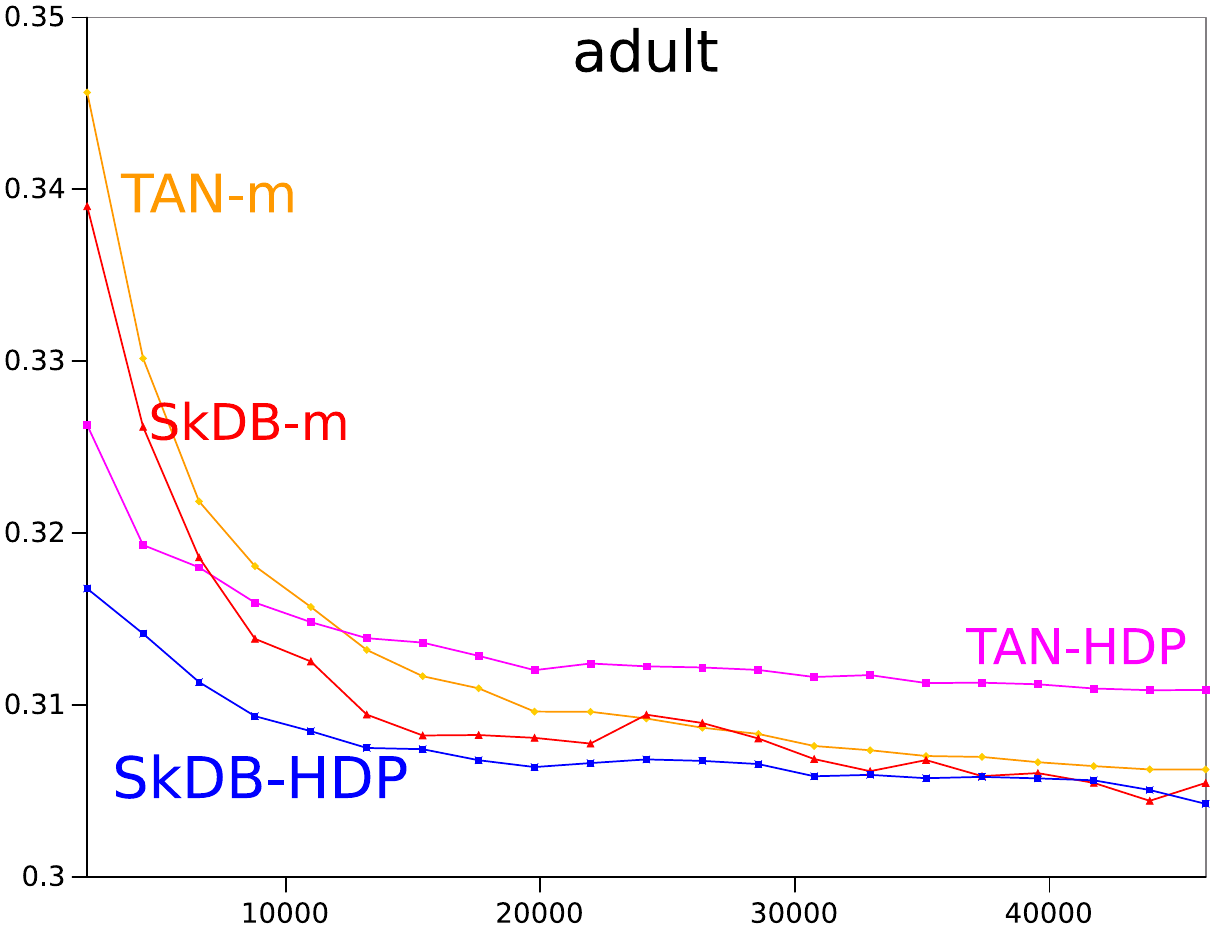}
\caption{\label{fig:learning-curves}Learning curves on RMSE for HDP and m-estimate.  The x-axis is dataset size, the y-axis is RMSE.}
\end{figure}

Finally, we present some learning curves for TAN and SkDB on a some larger datasets in Figure~\ref{fig:learning-curves}. Each point corresponds the mean RMSE for quantity of data $x$ over 10 runs. Globally, we can see that our HDP estimates seem to `learn' faster, i.e.\ overfit less. For the \texttt{connect-4} dataset, SkDB-HDP dominates all the way through with the difference in RMSE getting smaller as the quantity of data increases. For \texttt{adult}, we can observe the same behaviour for SkDB. Interestingly, for TAN on this dataset, although HDP estimates do learn faster, they are overtaken by m-estimates after 10,000 datapoints. 

\subsection{BNCs with HDP vs random forest}
\label{subsec:exp-BN-vs-RF}
Having shown that our approach outperforms the state of the art for BNCs parameter estimation, we compare BNCs using our HDP estimates against random forest (RF). 
The aim of this section is not to suggest that BNCs should replace RF, but rather that BNCs can perform competitively. 

Before proceeding, it is important to recall that RF is run on the same datasets as our BNCs with HDP estimates, i.e., with attributes discretized when necessary.

\begin{figure}
\centering
     \includegraphics[width=.5\linewidth]{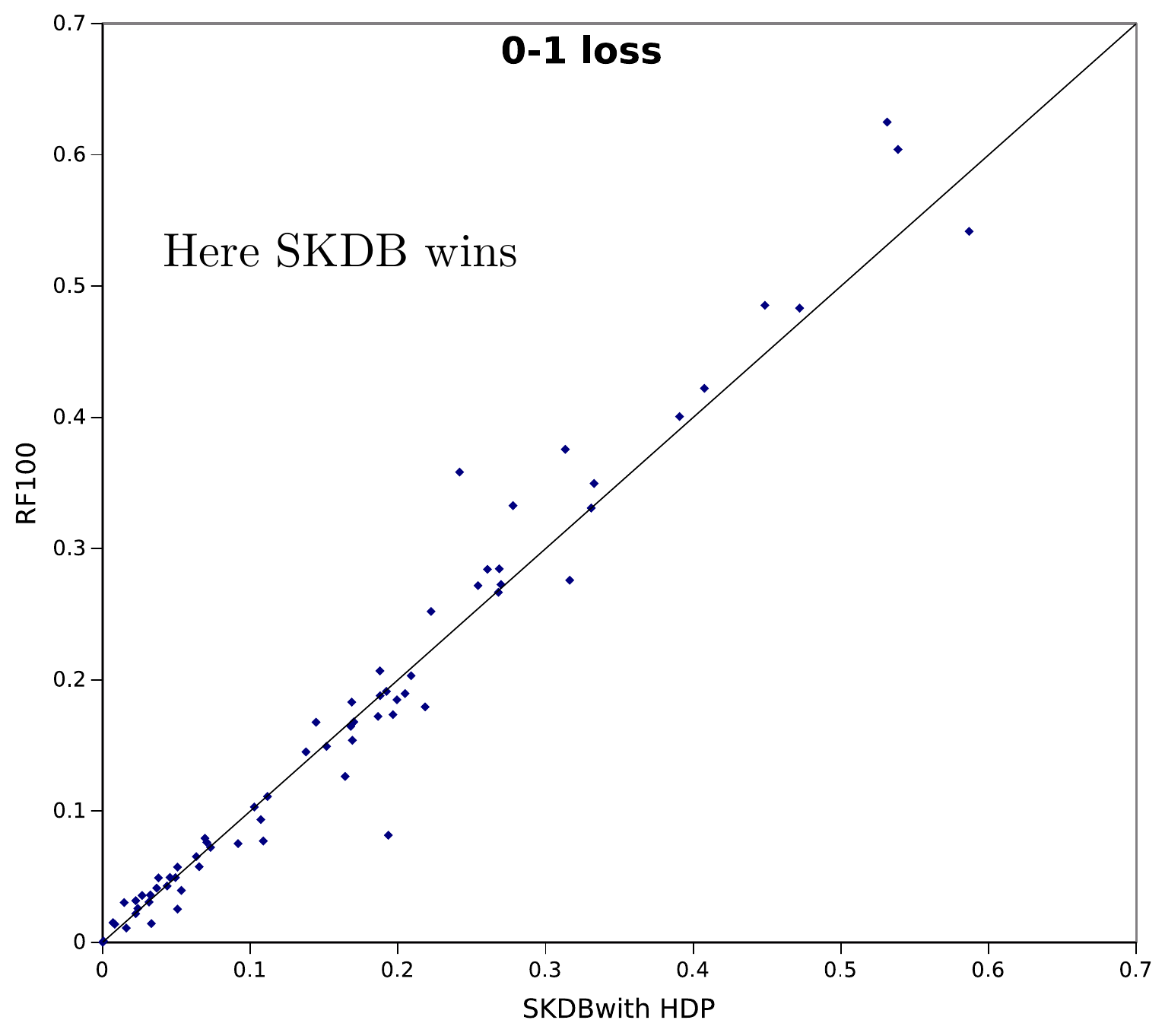}
    \caption{\label{fig:HDP_VS_RF_Scatter}0-1 loss scatter plot of SkDB with our HDP parameter estimate vs Random Forest}
\end{figure}

We report in Table~\ref{tab:BN-vs-RF} and Figure~\ref{fig:HDP_VS_RF_Scatter}
the results of TAN and SkDB. From this table we can see that RF is generally more accurate than the BNCs with m-estimates. Conversely, we can see that BNCs with HDP outperform RF more often, even with a model as simple as TAN.  This result is important because our techniques are all completely out-of-core and do not need to retain the data in main memory, as do most state-of-the-art learners. Note that comparing 0-1 loss is probably fairer to RF, because RF is not a probabilistic model (even if plain RF estimates as we do have been reported to outperform other RF variations in terms of RMSE \cite{bostrom2012forests}).

Obviously, for the larger datasets, RF catches up to TAN-HDP (which has a high-bias structure) but for the 10 largest datasets we considered, TAN-HDP still wins 6 times (1 draw) and SkDB-HDP is extremely competitive with a win-draw-loss of 7--0--3. 
\begin{table}
	\centering
	\caption{\label{tab:BN-vs-RF}Win/Draw/Loss m-estimates and our HDP estimates, as compared with Random Forest. We use our 2 flagship classifiers TAN and SkDB. Stat. sig. results ($p<0.05$) are depicted in boldface. }
	\begin{tabular}{lll}
    \hline \\[-5pt]
        Compared classifiers&\multicolumn{2}{l}{Win--draw--loss}\\[2pt]
		\cline{2-3} \\[-5pt]
		 &0/1-loss & RMSE \\[2pt]
		\hline \\[-5pt]
		TAN-$m$ vs RF & {26--3--39} & {\textbf{25--0--43}} \\[2pt]
		SkDB-$m$ vs RF & {27--3--38} & {29--1--38} \\[1pt]		\hline \\[-7pt]
        TAN-HDP vs RF & \textbf{42--3--23} & \textbf{42--0--26} \\[2pt]
		SkDB-HDP vs RF & 35--3--30 & \textbf{44--0--24} \\[2pt]
		\hline
	\end{tabular}
\end{table}

\subsection{Out-of-core capacity}
Our last set of experiments aims at showcasing the out-of-core capacity of our system. \modif{We run SkDB on the Splice dataset \cite{splice} --~which contains 50 million training examples and is provided with a test dataset with 5M samples~-- and compare our HDP estimates to the $m$-estimates. Note that this dataset is imbalanced with only 1\% of examples for the positive class. }

\modif{On this dataset, RF could not run using Weka defaults, requiring more than our limit of 138GB of RAM. We thus used instead XGBoost \cite{XGBoost}, which is the state of the art for scalable mixture of trees (here boosting) and used widely by data scientists to achieve state-of-the-art results on many machine learning challenges (XGBoost was used in 17 out of 29 winning solutions in the machine learning competition site Kaggle in 2015 \cite{XGBoost}). We use XGBoost's default parameters as per version 0.6 -- we use maximum depth of 6 and 50 rounds of boosting. Similarly to the previous, the aim of this section is not to suggest that BNCs should replace XGBoost, but rather to show that BNCs are an interesting set of models that can perform out-of-core and perform competitively when using our HDP-estimates. }

The results are reported in Table~\ref{tab:Splice}.
\modif{They show that HDP dramatically improves both 0-1 loss and RMSE as compared to $m$-estimates. 
Note that $m$-estimates would even be outperformed in terms of error-rate by simply predicting the majority class. Comparison with XGBoost is interesting, it shows that SkDB5 with our HDP estimates comes very close to XGBoost in terms of 0-1 loss. In terms of probability calibration our HDP estimates even push BNCs beyond XGBoost's performance, as evidenced by the RMSE. }

\begin{table}
  \centering
  \caption{\label{tab:Splice}Results on the Splice dataset on which RF cannot run. }
  \begin{tabular}{lll}
    \hline \\[-5pt]
    Classifier &0/1-loss& RMSE\\[2pt]
    \hline \\[-5pt]
    {SkDB5-$m$} & 1.499\% & 0.1093 \\[2pt]
    {SkDB5-HDP} &  0.318\% & 0.0544\\[2pt]
    \modif{XGBoost} & \modif{0.314\%} & \modif{0.0594} \\[2pt]
    \hline
  \end{tabular}
\end{table}

\subsection{Running time}
Although running time is not directly a focus of this paper, we give below some associated observations:
\begin{itemize}
 \item Training time complexity increases linearly with the number of iterations the sampler runs for, linearly with the number of covariates and linearly with the number of nodes in the trees (which increases exponentially with depth).
\item Training time is reasonable. As an example, training of {\it SkDB$_5$-HDP} (with $maxK=5$) on Splice with 50 million samples took under 4 hours, among which 1.5 hours are spent to learn the structure of the BN. {\it SkDB$_5$} implied that the 140 independent hierarchies have a depth of 6 and we run 5,000 iterations of the sampler. \modiff{This also implies that {\it SkDB$_5$-m} takes a bit more than 1.5 hours to be trained. XGBoost --~which is a highly optimised package~-- on Splice required just under one hour of computation.  }
\item For the Adult dataset training {\it SkDB$_5$} with 25k samples and 50,000 iterations with level tying took 86 seconds, for the Abalone dataset training with 2k samples took 6 seconds -- classification time takes less than 1s to classify 25k samples, which is one of the strength of BNCs: once learned, classification is a simple look-up for each factor. This classification time is actually under 1s for all models considered in this paper for the Adult dataset. 
\end{itemize}

\section{Conclusions}
\label{sec:Conclusions}
This paper presents accurate parameter estimation for
Bayesian network classifiers using hierarchical Dirichlet process estimates,
combining these well-researched areas for the first time.
We have demonstrated that HDPs are not only capable of
outperforming state-of-the-art parameter estimation techniques, but do so
while functioning completely out-of-core. We have also showed that,
for categorical data, this makes it possible to make BNCs highly competitive with random forest.
We note that while BNCs are not currently state of the art
for classification, they are still popular in applications.
With this improvement in performance, and usable implementations 
in packages such as R, BNCs will be far more useful in real-world applications
because they are readily implemented on high performance
desktops, and do not require a cluster.

This work naturally opens up a number of opportunities for future
research. First, we would like to perfect our sampler by assessing the
influence of the different runtime configurations
of our system including: how often should
we sample concentration, widening the window of pseudo-counts at the
start of the system and burn-in. Second, we would like to extend
this work to Pitman-Yor processes, which offer an exciting avenue for
research, in particular for variables with high cardinality.
Third, we would like to extend this framework to the general
class of Bayesian networks.



\begin{table}[bt] \center
  \begin{tabular}{p{3cm}p{8cm}}
    \hline
    Notation & Description \\
    \hline
    $n$		& Number of attributes -- also number of variables used to estimate the conditional probability\\
    $N$		& Number of data points in $\data$\\
    $Y$		& Random variable associated with class label -- also $X_0$ \\
    $y$		& value taken by $Y$ \\
    $|Y|$	& Number of classes\\
    $X_i$	& Random variable associated with attribute $i$ \\
    $x_i$	& value taken by $X_i$ \\
    $X_c$	& child variable for which we are estimating the conditional probability \\
    $\theta_{X_c|y,x_1,\cdots,x_n}$	& parameter vector associated with leaf node (at level $n+1$) for values $y,x_1,\cdots,x_n$ \\
    $\phi_{X_c|y,x_1,\cdots,x_i}$ & latent prior parameter for node at level $i$ associated with branching values $y,x_1,\cdots,x_i$\\
    $\alpha$ & concentration parameter for the Dirichlet distributions\\
    $n_{x_c|y,x_1,\cdots,x_n}$ & leaf-node parameter representing the number of data points with values $x_c|y,x_1,\cdots,x_n$\\
    $n_{x_c|y,x_1,\cdots,x_i}$ & intermediate-node parameter representing the number of data points received from its children nodes $n_{x_c|y,x_1,\cdots,x_{i-1}}=\sum_{x_i} t_{x_c|y,x_1,\cdots,x_i}$\\
    $t_{x_c|y,x_1,\cdots,x_i}$ & latent variable representing the fraction of $n_{x_c|y,x_1,\cdots,x_i}$ that is passed up to its parent\\
    $n_{.|y,x_1,\cdots,x_i}$ & marginal count $n_{.|y,x_1,\cdots,x_i}=\sum_{x_c}n_{x_c|y,x_1,\cdots,x_i}$\\
    $t_{.|y,x_1,\cdots,x_i}$ & marginal count $t_{.|y,x_1,\cdots,x_i}=\sum_{x_c}t_{x_c|y,x_1,\cdots,x_i}$\\
       \hline
\end{tabular}
\caption{\label{tab:notations}List of symbols used.}
\end{table}

\begin{table}
\centering
\small
\caption{Datasets}\label{UCIDatasets}
\setlength{\topmargin}{1pt} \setlength{\tabcolsep}{0.3pt}
\tabcolsep=1.0pt
\begin{tabular}{lrrrrlrrrrrlrrrrrr}
\hline
\bf Domain & \bf Case & \bf Att & \bf Class && \bf Domain & \bf Case & \bf Att & \bf Class  \\
\hline
Connect-4Opening&67557&43&3&&PimaIndiansDiabetes&768&9&2 \\
Statlog(Shuttle)&58000&10&7&&BreastCancer(Wisconsin)&699&10&2 \\
Adult&48842&15&2&&CreditScreening&690&16&2 \\
LetterRecognition&20000&17&26&&BalanceScale&625&5&3 \\
MAGICGammaTelescope&19020&11&2&&Syncon&600&61&6 \\
Nursery&12960&9&5&&Chess&551&40&2 \\
Sign&12546&9&3&&Cylinder&540&40&2 \\
PenDigits&10992&17&10&&Musk1&476&167&2 \\
Thyroid&9169&30&20&&HouseVotes84&435&17&2 \\
Mushrooms&8124&23&2&&HorseColic&368&22&2 \\
Musk2&6598&167&2&&Dermatology&366&35&6 \\
Satellite&6435&37&6&&Ionosphere&351&35&2 \\
OpticalDigits&5620&49&10&&LiverDisorders(Bupa)&345&7&2 \\
PageBlocksClassification&5473&11&5&&PrimaryTumor&339&18&22 \\
Wall-following&5456&25&4&&Haberman'sSurvival&306&4&2 \\
Nettalk(Phoneme)&5438&8&52&&HeartDisease(Cleveland)&303&14&2 \\
Waveform-5000&5000&41&3&&Hungarian&294&14&2 \\
Spambase&4601&58&2&&Audiology&226&70&24 \\
Abalone&4177&9&3&&New-Thyroid&215&6&3 \\
Hypothyroid(Garavan)&3772&30&4&&GlassIdentification&214&10&3 \\
Sick-euthyroid&3772&30&2&&SonarClassification&208&61&2 \\
King-rook-vs-king-pawn&3196&37&2&&AutoImports&205&26&7 \\
Splice-junctionGeneSequences&3190&62&3&&WineRecognition&178&14&3 \\
Segment&2310&20&7&&Hepatitis&155&20&2 \\
CarEvaluation&1728&8&4&&TeachingAssistantEvaluation&151&6&3 \\
Volcanoes&1520&4&4&&IrisClassification&150&5&3 \\
Yeast&1484&9&10&&Lymphography&148&19&4 \\
ContraceptiveMethodChoice&1473&10&3&&Echocardiogram&131&7&2 \\
German&1000&21&2&&PromoterGeneSequences&106&58&2 \\
LED&1000&8&10&&Zoo&101&17&7 \\
Vowel&990&14&11&&PostoperativePatient&90&9&3\\
Tic-Tac-ToeEndgame&958&10&2&&LaborNegotiations&57&17&2\\
Annealing&898&39&6&&LungCancer&32&57&3 \\
Vehicle&846&19&4&&Contact-lenses&24&5&3 \\
\hline
\end{tabular}
\end{table}

\section*{Compliance with Ethical Standards}
This work was supported by the Australian Research Council under awards DE170100037 and DP140100087.
The authors would like to thank Joan Capdevila Pujol and anonymous reviewers for helping us strengthen the original manuscript.

\bibliographystyle{spbasic}
\bibliography{biblio}

\end{document}